\documentclass[accepted]{uai2023} 

\usepackage[american]{babel}

\usepackage{natbib} 
\usepackage{mathtools} 
\usepackage{booktabs} 
\usepackage{tikz} 
\usepackage{amsfonts}       
\usepackage{graphicx}
\usepackage{subfig}
\usepackage{url}
\usepackage{rotating}

\DeclareMathOperator*{\argmin}{arg\,min}

\title{On the Informativeness of Supervision Signals}

\author[1]{\href{mailto:<is2961@princeton.edu>?Subject=Your UAI 2023 paper}{Ilia~Sucholutsky$^*$}}
\author[1]{Ruairidh~M.~Battleday$^*$}
\author[2]{Katherine~M.~Collins}
\author[3]{Raja~Marjieh}
\author[1]{Joshua~C.~Peterson}
\author[1]{Pulkit~Singh}
\author[2,4]{Umang~Bhatt}
\author[5]{Nori~Jacoby}
\author[2,4]{Adrian~Weller}
\author[1,2]{Thomas~L.~Griffiths}
\affil[1]{%
    Dept. of Computer Science\\
    Princeton University\\
}
\affil[2]{%
    Dept. of Engineering\\
    University of Cambridge\\
  }
  \affil[3]{%
    Dept. of Psychology\\
    Princeton University\\
}
  \affil[4]{%
    Alan Turing Institute\\
  }
\affil[5]{%
    Max Planck Institute for Empirical Aesthetics
  }
  
\begin{document}
\maketitle
\def\thefootnote{*}\footnotetext{Equal contribution.}


\begin{abstract}
  Supervised learning typically focuses on learning transferable representations from training examples annotated by humans. 
  While rich annotations (like soft labels) carry more information than sparse annotations (like hard labels), they are also more expensive to collect. For example, while hard labels only provide information about the closest class an object belongs to (e.g., ``this is a dog''), soft labels provide information about the object’s relationship with multiple classes (e.g., ``this is most likely a dog, but it could also be a wolf or a coyote'').
  We use information theory to compare how a number of commonly-used supervision signals contribute to representation-learning performance, as well as how their capacity is affected by factors such as the number of labels, classes, dimensions, and noise. 
  Our framework provides theoretical justification for using hard labels in the big-data regime, but richer supervision signals for few-shot learning and out-of-distribution generalization. We validate these results empirically in a series of experiments with over 1 million crowdsourced image annotations and conduct a cost-benefit analysis to establish a tradeoff curve that enables users to optimize the cost of supervising representation learning on their own datasets.

\end{abstract}

\section{INTRODUCTION}

Modern machine learning relies heavily on using large amounts of labeled data, and those labels typically come from annotations generated by humans. This raises an important question: how can we most efficiently use human annotations to create objective functions for machine learning systems? This is not just a matter of designing good interfaces and algorithms for collecting annotations---it involves a subtle interplay between the choices we make about the supervision signals we use to train our models and the difficulty of collecting the relevant annotations. For example,  soft labels for images (indicating uncertainty via a distribution over classes) are more expensive to collect than hard labels (indicating a single class), but are also potentially more informative to the learner ~\citep{peterson2019human,sucholutsky2021soft, Sucholutsky_Schonlau_2021,collins2022eliciting}. Making good choices about what questions to ask humans about our data requires understanding the informativeness of different supervision signals. 

In this paper, we explore this question for the case of {\em representation learning}, where the aim is to learn useful latent embeddings of input stimuli. Generally, training a neural network means learning successive layers of representations that will be used to perform some sort of task (e.g., classification).  The key decision in implementing a representation-learning framework often revolves around designing a supervision signal by quantifying the similarity between stimuli. Significant work has gone into the design of supervision signals for deep representation learning resulting in contrastive objectives \citep{chen2020simple,khosla2020supervised}, classification objectives \citep{huh2016makes,ridnik2021imagenet}, reconstruction objectives \citep{devlin2018bert, kingma2013auto}, and many others \citep{guo2019deep}.
For example, recent work has shown that models trained with hard labels on classification tasks can approximate the structure of human latent representations 
at a fraction of the cost of exhaustively collecting the pairwise-similarity judgments required for soft contrastive learning~\citep{peterson2018evaluating,marjieh2022words}. 
Thus, in many cases, multiple representation learning objectives could work, and it is not clear when one objective should be preferred. 


\begin{figure}[t!]
    \centering
    \includegraphics[width=0.87\linewidth]{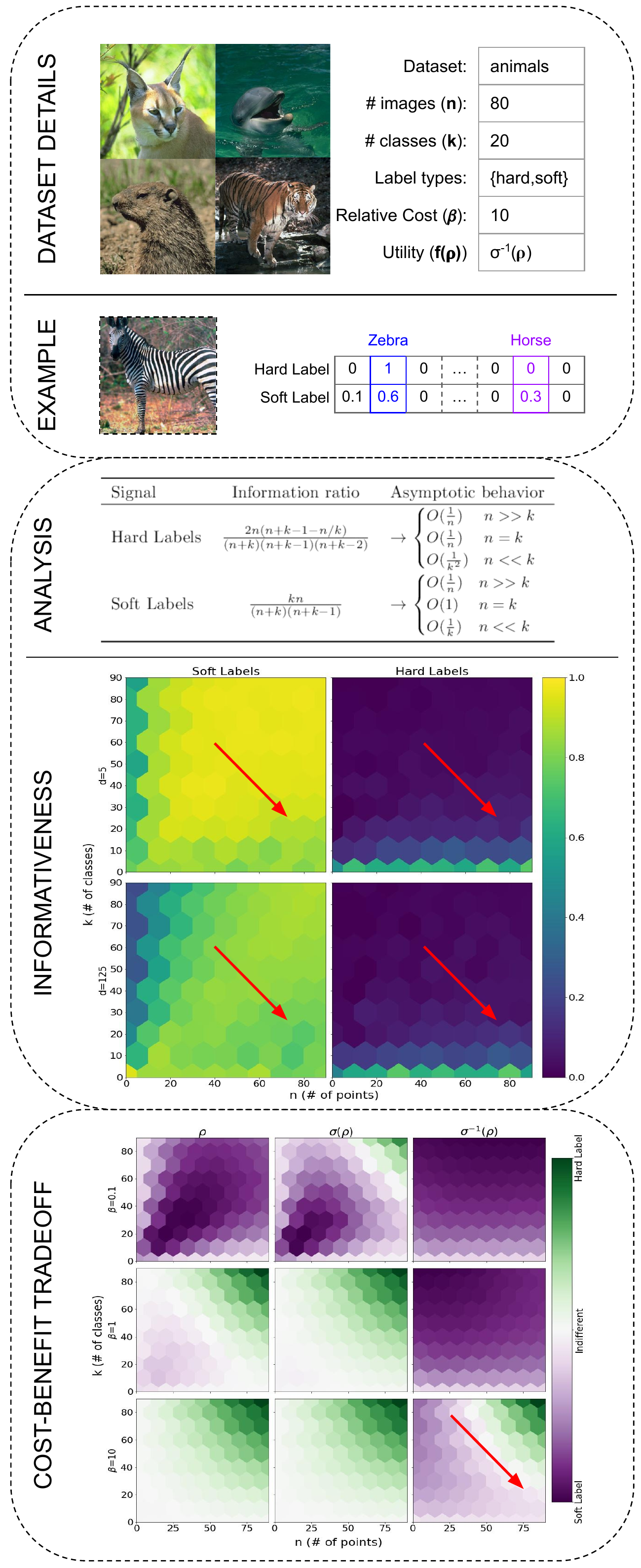}
    \caption{An example workflow for using our framework to decide what label type(s) to collect when annotating a dataset. \textbf{Top}: User specifies the task. \textbf{Middle}: Information-theoretic characterization of the task (e.g., many classes, but few examples). Heatmaps show correlation ($\rho$) of ground-truth similarities with pairwise similarities recovered from simulated labels when varying latent dimensionality ($d$), number of points ($n$), and number of classes ($k$). \textbf{Bottom}: Cost-benefit analysis of signal type based on subjective utility ($u(\rho)$), cost ($\beta$), number of points ($n$), and number of classes ($k$). Red arrows point to cells corresponding to the user-specified dataset.}
    \vspace{-2mm}
    \label{fig:sims}
\end{figure}

Our goal in this paper is to compare the informativeness of various supervision signals (i.e. types of annotations) to \textbf{empower researchers to optimize data annotation for their supervised representation learning tasks} (see Figure~\ref{fig:sims}). We \textbf{develop an information-theoretic framework (Section~\ref{sec:info_formalisms})} for analyzing supervision signals and use it to compare two popular supervision signals from  the classification literature---hard labels and soft labels. We quantify their relative (representational) information content by comparing them to similarity triplets (i.e., queries of the form ``Is $x$ more similar to $y$ than to $z$?''), a  supervision signal used in contrastive learning and cognitive science~\citep{jamieson2011low,hoffer2015deep}. We relate the information each signal provides to three common features of machine learning datasets: number of labels, number of classes, and dimensionality, and find that while both hard and soft labels provide information about hidden representations, their responses to these three variables are very different. \textbf{Simulations confirm these results (Section~\ref{sec:simulations}),} showing how the marginal information provided by each label translates into better representation learning performance. A \textbf{cost-benefit analysis (Section~\ref{sec:costbenefit})} comparing soft and hard labels shows there are meaningful differences between them: soft labels are expensive to collect but provide a considerable amount of representational information while hard labels are cheap but fairly uninformative (Figure~\ref{fig:sims}). To close this gap, we consider several types of sparse representations that enable selective interpolation between the soft- and hard-label regimes (Figures~\ref{fig:fig2},~\ref{fig:fig3}). We extend our analysis with these sparse supervision signals to \textbf{establish a tradeoff curve allowing users to optimize the cost of labeling their datasets for supervised representation learning (Section~\ref{sec:labelopt})}. 
Finally, we confirm the theoretical and simulation results by \textbf{human experiments (Section~\ref{sec:experiments})} with crowdsourced similarity judgments and various types of soft and hard labels collected for \texttt{CIFAR-10}~\citep{krizhevsky2009learning}. 


\section{Background}
\label{sec:signal_types}
Throughout this paper, we discuss a variety of supervision signals. In this section, we identify the practical settings these signals correspond to and summarize related work. 

\textbf{Pairwise similarity judgments}: 
For every pair of points in the dataset labelers are asked to rate the similarity on a fixed, bounded scale (e.g., a Likert scale). This signal has a long history of being employed by cognitive scientists to learn about (hidden) human representations of stimuli, typically by using an embedding method like multi-dimensional scaling (MDS)~\citep{shepard1980multidimensional}, and more recently can be found in state-of-the-art contrastive learning objectives~\citep{chen2020simple, khosla2020supervised}. 

\textbf{Triplet similarity judgments}:
For every set of three points $x,y,z$ in the dataset, labelers are asked to respond to queries of the form ``Is $x$ closer to $y$ than to $z$?'' This non-metric signal has also been used  as a method for learning human representations, often by applying embedding techniques like non-metric MDS~\citep{jamieson2011low,pmlr-v2-agarwal07a,6713995}; it is considered to be a more accurate alternative to pairwise similarity judgments as it is easier to compare two lengths than to provide consistent judgments over un-normalized scales. This signal is used in machine learning (typically with the triplet loss) and cognitive science~\citep{hoffer2015deep,hermans2017defense,hebart2020revealing,roads2021enriching}.


\textbf{Hard labels}: For each point, labelers  pick a single, most relevant class out of the fixed set of all classes in the dataset. This signal is typically used for classification, though it can also be used for representation learning via pre-training~\citep{huh2016makes,ridnik2021imagenet}.

\textbf{Soft labels}: For each point, labelers assign proximity or probability to each of the fixed set of all classes in the dataset. This signal is also used for classification and representation learning via pre-training, and can be more effective than hard labels, particularly in settings with small data~\citep{Xie_2020_CVPR, Sucholutsky_Schonlau_2021, sucholutskv2021one, Liu2021LAST, malaviya_sucholutsky_oktar_griffiths_2022}.

\textbf{Top-class soft labels}: The researcher picks a subset of size $\hat k$ of the classes in the dataset that maximize mutual information (estimated based on the already collected subset). For each point, labelers assign proximity or probability to each of  the fixed set of classes in the subset.  The number of classes can be reduced by simply taking an arbitrary subset, or more systematically via methods like label coarsening~\citep{hanneman2011automatic}. 

\textbf{Sparse soft labels}: For each point, labelers assign proximity or probability to each of the exactly $\hat k$ most relevant classes out of the fixed set of all classes in the dataset~\citep{collins2022eliciting}. In addition to the connections discussed above for the other soft label variants, this particular variant is also connected to top-k classification~\citep{lapin2017analysis} and soft vector quantization~\citep{seo2003soft}. 



\section{RELATIVE INFORMATIVENESS OF SUPERVISION SIGNALS}\label{sec:info_formalisms}
\textbf{Setup}:
We consider the scenario where a researcher collects an initial set of labels for a small subset of their dataset and wants to use it to optimize the labelling of their entire dataset (Figure~\ref{fig:sims}, top). The researcher has a number of options for what kind of labels to collect and wishes to maximize representation learning performance while minimizing labelling cost.
We formalize representation learning as the process of recovering a hidden (low-dimensional) latent structure from a set of (high-dimensional) stimuli (e.g. images). In particular, we focus on the non-metric setting where we want to recover the correct rank order of pairwise distances between all latent vectors. Our goal is to determine which supervision signal is most effective (in terms of both performance and cost) for representation learning.

\textbf{Problem Definition}:
Consider a set of stimuli $\{x_i\}_{i=1}^n \in \mathbb{R}^d$ with some associated latent representations $\{z_i\}_{i=1}^n \in \mathbb{R}^h$. The distance between each pair of latent vectors induces a relational order over latent pairs and our goal is to find a function $f:\mathbb{R}^d \rightarrow \mathbb{R}^h$ (where typically $h<<d$) such that it preserves this relational order, that is, $||f(x_i)-f(x_j)||\leq||f(x_i)-f(x_k)||$ iff $||z_i-z_j||\leq||z_i-z_k||$. Crucially, the latent vectors are accessible only implicitly via different supervision signals (or queries) such as hard and soft labels. We operationalize the informativeness of different supervision signals as the number of relational constraints that a naive learner can recover based on them (i.e., a learner that attempts to follow the signals as is without applying other geometric constraints such as triangle inequalities).  

\textbf{Triplet Constraints}:
Conceptually, when training a neural network for classification, providing a label for a point roughly corresponds to requiring that the network weights should be updated such that the embedding of this point will be closer to one class than to other classes. For this analysis, we assume that each class can be represented by its centroid (e.g., each class is unimodal), and so classification labels provide information about proximity of latent vectors to these centroids\footnote{This can easily be generalized to multi-modal classes by treating them as compositions of multiple unimodal subclasses (i.e. for an $m$-modal class, $m$ centroids need to be learned).}. When training with batches, providing labels for a batch additionally corresponds to requiring that the centroid of each class be closer to its associated set of embeddings than to the other embeddings in the batch. In both cases, neural networks are optimizing constraints of the form ``$x$ is closer to $y$ than to $z$,'' which we call ``triplet constraints''. We now formalize this concept to use it as a measure of information content in labels.

Suppose we have a system with $n$ labels, $k$ classes with centroids $C_1,...,C_k$, and  stimuli $\{x_i\}_{i=1}^n$ with latent representations $\{z_i\}_{i=1}^n$. 
A triplet constraint is an inequality of the form $||z_a-z_b||<||z_a-z_c||$. This can be rewritten as the query $r_{i, j, k}=\left\{x \in \mathbb{R}^{d}:\left\|f(x_{j})-f(x_{i})\right\|<\left\|f(x_{k})-f(x_{i})\right\|\right\}$, and each such query provides at most one bit of information as it cuts in half the space where $x_i$ can be located ~\citep{jamieson2011low}. 
For any set of three stimuli, there are three unique queries: $r_{i,j,k}, r_{j,i,k}, r_{k,i,j}$. Thus, the total number of unique queries for $n$ stimuli is $3{n \choose 3}$. 
However, in the case of hard and soft labels, we make queries not only in terms of the $n$ objects but also in terms of the $k$ class centroids. In other words, we are seeking to recover embeddings not only for the $n$ points of interest, but also $k$ additional reference embeddings. As a result, the total number of unique queries in these cases is $3{n+k \choose 3}$. 

\textbf{Hard Labels}:
We define the hard label for stimulus $x$ as a vector $l$ of length $k$, such that $l_i=
1 \text{ if } i=\argmin_j(||f(x)-f(C_j)||) \text{ and } 0 \text{ otherwise}.$
We can now extract two types of triplet queries. The first is a triplet consisting of a class centroid $C_i$, a stimulus $x_P$ that is a ``positive'' example of this class, and a negative example stimulus $x_N$. The query has the form $||f(x_P)-f(C_i)||<||f(x_N)-f(C_i)||$. The second is a triplet consisting of a stimulus $x_i$, the class centroid that is closest to it ($C_P$), and another class centroid further away ($C_N$). This query has the form $||f(C_P)-f(x_i)||<||f(C_N)-f(x_i)||$. If the hard labels are distributed evenly between $k$ classes, then on average there are $n/k$ stimuli per class. Then $n$ hard labels give us $n(k-1)$ constraints of the second type and $k(n/k)(n-n/k)=n^2(1-1/k)$ of the first type---a total of $T_H(n,k)=n(k-1)+n^2(1-1/k)$  constraints. 

\textbf{Soft Labels}:
To produce a probability distribution over classes, neural networks often have a softmax activation function after the output layer~\citep{bridle1989training,martins2016softmax,krizhevsky2017imagenet}.  Accordingly, we define the soft label for a stimulus $x$ as a vector $l$ of length $k$, such that $l_i=\frac{e^{-||f(x)-f(C_i)||}}{\sum_je^{-||f(x)-f(C_j)||}}$. 
There are again two types of triplet queries that we can extract from soft labels. The first is a triplet of the form $||f(x_P)-f(C_i)||<||f(x_N)-f(C_i)||$ where $C_i$ is the centroid of class $i$ and $x_P, x_N$ are two training set points with corresponding soft labels $l^P, l^N$ such that $l^P_i > l^N_i$.
The second  is a triplet consisting of the form $||f(x)-f(C_i)||<||f(x)-f(C_j)||$ where $x$ is a training set point corresponding to label $l$ and $C_i, C_j$ are the centroids of classes $i,j$ such that $l_i>l_j$. Our $n$ soft labels thus give us $nk(k-1)/2$ constraints of the second type and $kn(n-1)/2$ of the first---a total of $T_S(n,k)=kn(k+n-2)/2$ triplet constraints. 


\textbf{Information Ratio}:
While we now have a measure of how much information each label provides, it is unclear how much information is actually needed to recover human-aligned representations for all the objects. Intuitively, we would expect that more information is required when more objects are being embedded (i.e. when $n+k$ increases). We can normalize our results from the previous section to account for this by taking the ratio of the number of constraints we can recover from a set of labels to the total number of possible queries (i.e. $IR(n,k)=\frac{T(n,k)}{3{n+k \choose 3}}$). This ``information ratio'' may be a proxy for how much information we are recovering about the latent representations. We present the information ratios for hard and soft labels in the ``Analysis'' portion of Figure~\ref{fig:sims} along with their asymptotic behavior in three regimes: the many-shot case (where there are many more points than classes), the one-shot case (where there is one point per class; \citep{1597116}), and the less-than-one-shot case (where there are fewer points than classes; \citep{Sucholutsky_Schonlau_2021}). 

Our results predict three scaling phases for soft labels and two scaling phases for hard labels. In particular, hard labels and soft labels are predicted to have similar asymptotic behavior in the many-shot case which may explain why pre-training on very large datasets using hard-label classification is effective (e.g. \citep{huh2016makes,ridnik2021imagenet}). However, soft labels are predicted to have much better representation learning performance in the one-shot and less-than-one-shot cases. Notably, the results predict that in the one-shot case, the quality of representations learned from soft labels should not degrade (as it does in every other case) when the number of points and classes increases.

\textbf{Representation Learning as Communication}: Our representation learning setup can be seen as a communication problem where we want to recover a fixed number of bits about hidden representations of a black-box model. The model can be queried via multiple channels, each of which has its own encoder and corresponds to a different choice of supervision signal. As established by~\cite{jamieson2011low}, each triplet query is a single bit. In information theory, the efficiency, also known as the normalized entropy, of each channel is defined as $\eta(X) = \frac{H(X)}{H_{max}(X)}$ where $H_{max}$ is max entropy (i.e. the total number of bits) and $H$ is entropy (i.e. the number of bits remaining with unknown states after transmission). We defined information ratio as the ratio of the number of bits recovered to the total number of bits, and thus a stochastic version of information ratio (with data sampled from a distribution instead of being fixed) can be defined as $IR(X)=\frac{H_{max}(X)-H(X)}{H_{max}(x)}=1-\eta(X)$.

\textbf{Signal-to-Noise Ratio}: Our triplet analysis implicitly assumed that the noise distribution was the same between each type of label and could thus be ignored when comparing their relative informativeness. However, in practice, eliciting different kinds of annotations may be associated with different levels of noise due to changes in difficulty, user interface, participant pools, etc. Within our information-theoretic framework, noise can be viewed as the probability $\epsilon_s$ of a bit flip (i.e. that we get the incorrect response to a query of the form ``Is $x$ closer to $y$ than to $z$?'') during communication over channel $s$ . If a set of $n$ labels (with $k$ classes) of type $s$ provides $T_s(n,k)$ triplet constraints under our framework in the noise-free case, then the number of constraints in the noisy case is just $(1-\epsilon_s)T_s(n,k)$ where $\epsilon_s$ is the bit flip rate for labels of type $s$. 






\begin{figure*}[htb!]
    \centering
    \includegraphics[width=0.25\textwidth]{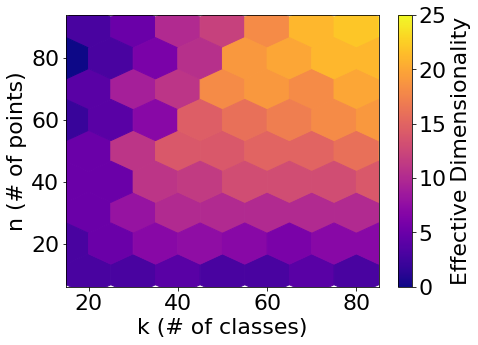}
    \includegraphics[width=0.74\textwidth]{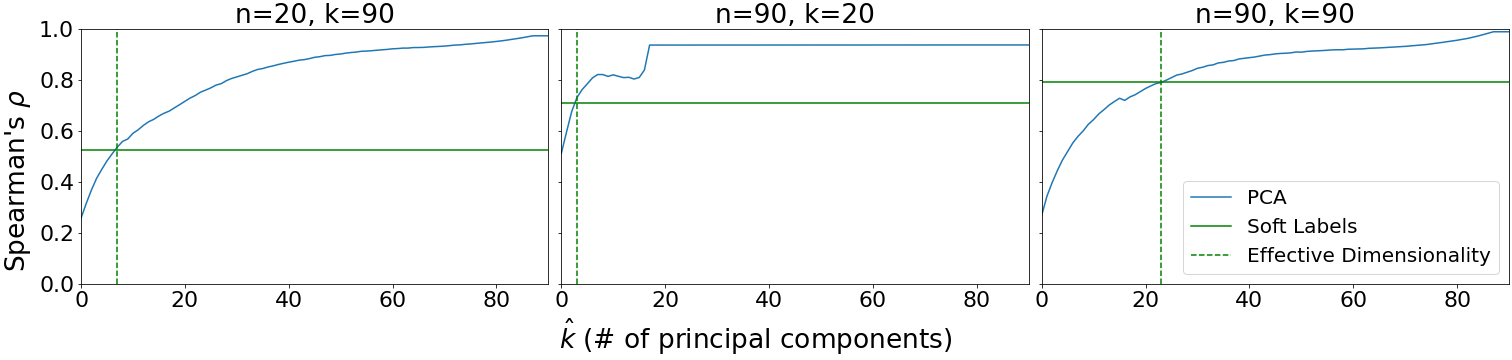}
    \vspace{-1mm}
    \caption{\textbf{Left}: Effective dimensionality of soft labels at different combinations of $n$ and $k$. \textbf{Right}: Three examples of PCA curves used to compute effective dimensionality of soft labels for every combination of $n$ and $k$.}
    \vspace{-4mm}
    \label{fig:fig2}
\end{figure*}
\section{SIMULATIONS}\label{sec:simulations}
Our analysis predicts soft labels should generally lead to better performance than hard labels, particularly when there are few labels and many classes. However, we still need to understand how information ratios actually translate to representation learning performance. We conducted simulations to see the effect of four variables (and their interactions) on performance: label type (soft or hard), number of points ($n$), number of classes ($k$), and latent dimension ($d$).
We consider values of $n$ and $k$ in the range of $[3,90]$, and $d\in\{5,25,125\}$. For each combination of $n,k,d$ we sample a total of $n$ points from Gaussians centered at $k$ random locations $C_1,...,C_k \in \mathbb{R}^d$. 
We computed hard and soft labels for these points using the equations defined above and then mine all triplet constraints of both types from both sets of labels.  We apply Generalized Nonmetric Multi-Dimensional Scaling (GNMDS;  \cite{6713995}) to both sets of queries to find embeddings that best fit the respective triplet constraints. The Gram matrix outputted by GNMDS can be interpreted as predicted (unnormalized) pairwise cosine similarities between all $n+k$ points and centroids.

To understand how much information we recover from each of these two sets of queries, we construct a matrix of the true pairwise cosine similarities for the set of all $n+k$ points and class centroids and compute the Spearman rank correlation ($\rho$) between the upper triangle of the Gram matrices and the ground truth matrix. Thus, a higher $\rho$ corresponds to better recovery of the underlying latent representations. We visualize the simulation results in Figure~\ref{fig:sims}. The results confirm the theoretical findings from the previous section. Specifically, the three phases for soft labels and two phases for hard labels match our analytical results, and \textbf{a higher information ratio translates into better performance}.

\section{COST-BENEFIT TRADEOFFS}\label{sec:costbenefit}
We can now construct cost-benefit tradeoff curves to \textbf{determine when a user would prefer to use one signal over the other}. Suppose we define $\rho$ as above, and subjective utility as $U(\rho)$. This utility function can take many forms (e.g., $U(\rho)=b\rho \text{ or } b\sigma(\rho) \text{ where } \sigma \text{ is the sigmoid function and } b>0$). If we assume that the cost of collecting a soft label over $k$ classes is about $k$ times more expensive than collecting a hard label, we can define the subjective loss function as $L_s=C(s) - U(\rho), \text{ where } C(s)=cn \text{ if } s\in S_{hard} \text{ and } cnk \text{ if } s\in S_{soft}$.
This is equivalent to optimizing $\hat L = \frac{c}{b}\hat c(s) - \hat u(\rho)$ which we can re-parametrize to a form reminiscent of the information bottleneck~\citep{tishby2000information}: $\hat L = \beta\hat c(s) - \hat u(\rho)$. Since we have shown that the information ratio, which we define as $\hat\rho$, can provide us with an estimate for $\rho$, we can also replace $U(\rho)$ by $U(\hat\rho)$. 

We investigate cost-benefit tradeoffs by varying $(\beta, \hat u)$, showing the results for several combinations in Figure~\ref{fig:sims}. While the results depend greatly on choice of $\hat u$, a few regularities emerge. First, in all cases, regardless of $\beta$, when the number of classes ($k$) or the number of points ($n$) is low, soft labels are roughly as preferred as, or more preferred, than hard labels. Second, when there is an emphasis on cost (i.e. high $\beta$), hard labels become preferable as $n$ and $k$ both increase, but when the emphasis is on performance (i.e. low $\beta$), soft labels remain preferable as $n$ and $k$ increase.

\section{LABEL OPTIMIZATION}\label{sec:labelopt}

\begin{figure}[t!]
    \centering
    \includegraphics[width=\linewidth]{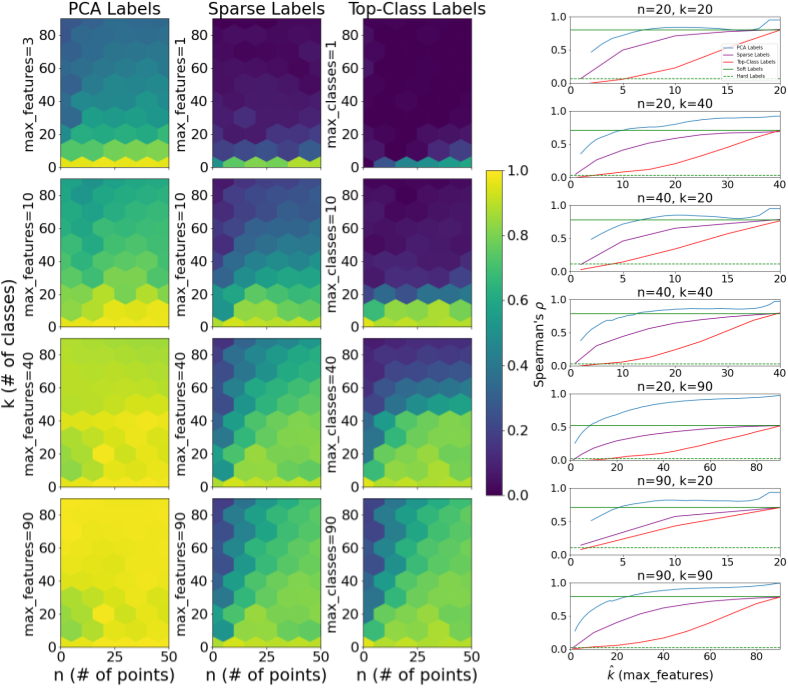}
    \caption{Label sparsity. \textbf{Left}: Spearman rank correlation ($\rho$) of pairwise similarities recovered from PCA labels, sparse labels, and top-class labels when varying the maximum number of features/classes in the labels ($\hat k$), number of points in the dataset ($n$), and number of classes in the dataset ($k$). \textbf{Right}: Comparison of sparsity curves for PCA labels (blue), sparse labels (purple), and top-class labels (red) for several combinations of $n$ and $k$. Straight lines correspond to soft label (solid green) and hard label (dashed green) performance. }
    \label{fig:fig3}
    \vspace{-4mm}
\end{figure}
\textbf{Effective Dimensionality}:
While soft labels appear to be more informative than hard labels, they are not necessarily an optimal encoding for efficiently communicating information about representations. In order to study how an optimal encoding might perform, we run principal component analysis (PCA) on our simulated datasets and observe the effects of varying the number of PCs that are retained ($\hat k$). The resulting performance curves provide an approximate upper bound on how well any set of vectors of length $\hat k$ (e.g. a set of soft labels) can communicate information about representations. We can also use these curves to understand how efficient soft labels are at this task. Specifically, we define the ``effective dimensionality'' of a set of soft labels as the minimum number of PCs ($\hat k$) necessary to achieve the same representation learning performance as when using the soft labels in the way we described in the previous section. In Figure~\ref{fig:fig2}, we show the effective dimensionality of soft labels for different combinations of $n$ and $k$ along with several examples of the PCA curves. We found a strong positive correlation between information ratio and effective dimensionality ($r=0.734, p<10^{-15}$)  providing further evidence information ratio is a useful metric for evaluating representation learning signals and predicting performance.

\textbf{Label Sparsity}:
Since the effective dimensionality of soft labels appears to generally be much lower than the number of components of each soft label ($k$), this suggests that soft labels are an inefficient encoding for representational information, potentially due to redundancy. We now investigate two methods for remedying this inefficiency by introducing sparsity into soft labels in a disciplined manner. 

The first method, which we call ``top-class soft labels'', introduces sparsity globally by ignoring classes that are the least informative across the entire dataset. Formally, we construct a matrix $X$ where the $i$-th row correspond to the $i$-th soft label and the $j$-th column corresponds to the $j$-th class, and estimate the mutual information between each column and the ground-truth similarity matrix. We then keep only the top $\hat k$ most informative columns and set the rest to $0$. 

The second method, which we call ``sparse soft labels'', is to introduce sparsity locally by ignoring classes that are least informative for each point. Formally, we again construct a matrix $X$ as above, but now for each row, we individually keep only the $\hat k$ largest components and set the rest to $0$. These two methods provide a way to reduce the cost of collecting soft labels, while retaining much of the information. 

In Figure~\ref{fig:fig3}, we visualize representation learning performance when using PCA, sparse labels, or top-class labels at different levels of sparsity. While sparse labels consistently outperform top-class labels (which is to be expected since sparse labels provide finer control over how sparsity is induced), we note that there is also a gap between PCA and sparse labels. This again suggests that it may be possible to design more effective labeling methods than soft labels, potentially by defining a search space over possible classes and applying a procedure like PCA to optimize over it. We leave this as a promising direction for future work.

\textbf{Optimizing Label Collection}:
Since top-class and sparse labels provide a way to selectively interpolate between the soft label and hard label regime, we can now use sparsity to optimize the cost-benefit tradeoff curves discussed above beyond the binary preference optimization shown at the bottom of Figure~\ref{fig:sims}. We use the same loss functions as above with several combinations of cost parameter $\beta$ and utility function $\mu$, but we now apply them to top-class and sparse labels at various levels of sparsity ($\hat k)$. We visualize a number of these cost-benefit tradeoff curves in the Supplement. By picking the $\hat k$ that corresponds to minimal loss, we can now optimize our label collection to minimize cost while maximizing performance. The results suggest that \textbf{using sparse labels and picking the right level of sparsity ($\hat{k}$) can often provide big gains} as opposed to using either hard labels or regular (dense) soft labels.



\section{Experiments}\label{sec:experiments}
In this section, we investigate how our theory applies to real soft labels crowdsourced from human annotators. First, we present several methods for collecting soft labels---including new sets of annotations that we crowdsource for this study---and place them into the framework described above. Second, we use a large dataset of similarity judgements to assess the informativeness of the different label types. Finally, we assess how the inductive biases conferred by different label types affect the classification performance of a range of convolutional neural networks (CNNs).



\subsection{Experimental Setup}

We consider a range of supervision signals over the \textit{testing} subset of \texttt{CIFAR-10} images \citep{krizhevsky2009learning}, including hard labels, smoothed hard labels, soft labels, and similarity judgments. Each label type represents a different supervision signal presented in Section~\ref{sec:signal_types}. Each type of soft label was collected using different experimental interfaces, details of which are in the Supplement.


\textbf{CIFAR-10H}: The \texttt{CIFAR-10H} labels, originally collected by \cite{peterson2019human, battleday2020capturing}, are derived by averaging over 500,000 crowdsourced hard labels (roughly 50 per image). These are then normalized at the image level to return probability distributions. 

\textbf{CIFAR-10DS}: The \texttt{CIFAR-10DS} labels, made up of over 500,000 judgments we crowdsourced for this study, are \textit{dense} soft labels (i.e., assigned over all classes in \texttt{CIFAR-10}). Annotators provided numerical judgments on a 0 (not at all) to 1 (completely) scale using sliders depending on how well each category described a certain image. 

\textbf{CIFAR-10S}: The \texttt{CIFAR-10S} labels from \citeauthor{collins2022eliciting} are \textit{sparse soft labels}. Annotators provided around 20,000 judgments about the likelihood of the top two categories for each image and any categories which they believed were definitely wrong (referred to as a ``clamp'', and assigned zero probability). As the authors only collected such labels over 1,000 examples from the test set, we in-fill the remaining 9,000 with either: 1) hard labels (\texttt{CIFAR-10S+hard}) or 2) simulated top-2 soft labels (\texttt{CIFAR-10S+dense}). 

\textbf{CIFAR-10T} The \texttt{CIFAR-10T} labels are a novel set of labels we crowdsourced, comprising over 350,000 typicality ratings for each image under the ground truth category (about 35  judgements per image). 1759 unique participants were recruited on Amazon Mechanical Turk, and presented with a sequence of 200 randomly sampled \texttt{CIFAR-10} test set images, upsampled to 160x160 pixels (see \cite{peterson2019human, battleday2020capturing}. Participants were given the category of each image, and asked to rate how
typical it was of the category on a sliding scale
of ``Not at all typical" to ``Extremely typical". We interpret an image's typicality as the probability of the ground truth class, and spread the remaining probability mass over the 9 remaining labels---a smoothed version of a \textit{sparse} soft label with K=1).

\textbf{CIFAR-10LS}: We derive two more sets of soft labels by applying label smoothing (LS) to the \texttt{CIFAR-10} hard labels. Unlike human-derived soft labels, the ``softness'' here is applied uniformly and independently of the associated image. This allows us to control for the previously observed regularization effects of label smoothing~\citep{muller2019labelSmoothHelp}. We pick the smoothing rate, $\epsilon$, to roughly match the distributions of our crowdsourced labels. The ``low'' level, $\epsilon\approx0.05$, is the average probability mass per soft label for the 9 non-maximal categories in the \texttt{CIFAR-10H} dataset. The ``high'' level, $\epsilon\approx0.2$, matches the \texttt{CIFAR-10DS} dataset.


\textbf{Similarity Judgments}: We elicited a total of 49,500 pairwise similarity judgments over two subsets each consisting of 100 \texttt{CIFAR-10} test set images by having human annotators rate the similarity between unlabeled image pairs on a Likert-scale ranging from 0 (completely dissimilar) to 6 (completely similar). The images were deliberately chosen to be ambiguous about class (see Supplement).

\subsection{Label Informativeness}

\textbf{GNMDS Results}:
We first repeat our GNMDS analyses from the theory and simulation sections on all the \texttt{CIFAR-10} label variants for 200 high-entropy images (see Supplement). 
We use GNMDS to get triplet-respecting embeddings for each label type and then compute several metrics. As before, we compute Spearman correlations ($\rho$) between the \texttt{CIFAR-10} GNMDS and elicited similarity judgments. Our previous analyses assumed a consistent signal-to-noise ratio, or error rate, across all the label types. However, with our set of CIFAR-10 label variants coming from different participant pools and elicitation pipelines, it is likely that error rates will vary between label types. We approximate error rates of each label type by counting the proportion of triplet queries derived from the GNMDS embedding whose binary responses do not match the responses from the corresponding triplets computed from the human similarity judgments (i.e. the bit flip rate). Additionally, we compute label entropy and the variance of the first order statistic (i.e. variance of the probability mass assigned to the class with the highest assigned probability) for each label variant. 

\begin{figure}[t!]
    \centering
    \includegraphics[width=0.45\textwidth]{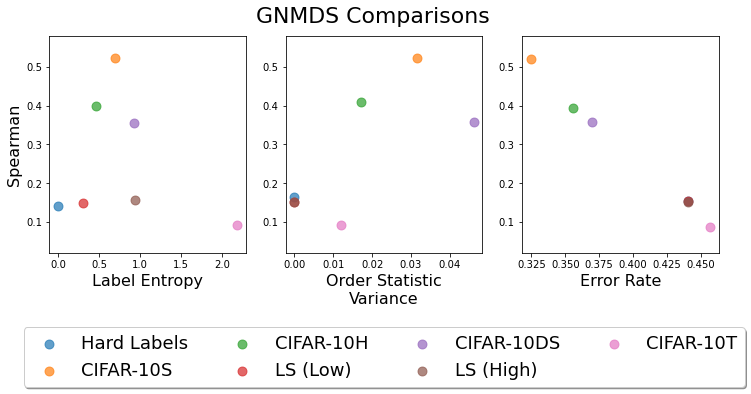}
    \caption{Correspondence between various metrics and Spearman rank correlation for each label variant.}
    \vspace{-5mm}
    \label{fig:gnmdsHuman}
\end{figure}

We visualize these results in Figure~\ref{fig:gnmdsHuman}. We find that representation learning performance (measured by Spearman correlation) generally increases as softness (measured by entropy and variance) increases, but does not increase when smoothing hard labels. However, \textbf{we find there is a ``sweet spot'' for softness} after which performance begins to decrease. We also find the expected linear relationship between error rate and performance that our framework predicts. We hypothesize that the U-shaped relationship between softness and performance may partially be caused by increasing error rates, and partially by a resonance effect (see Supplement).

\begin{figure*}[t!]
    \centering
    \includegraphics[width=0.9\textwidth]{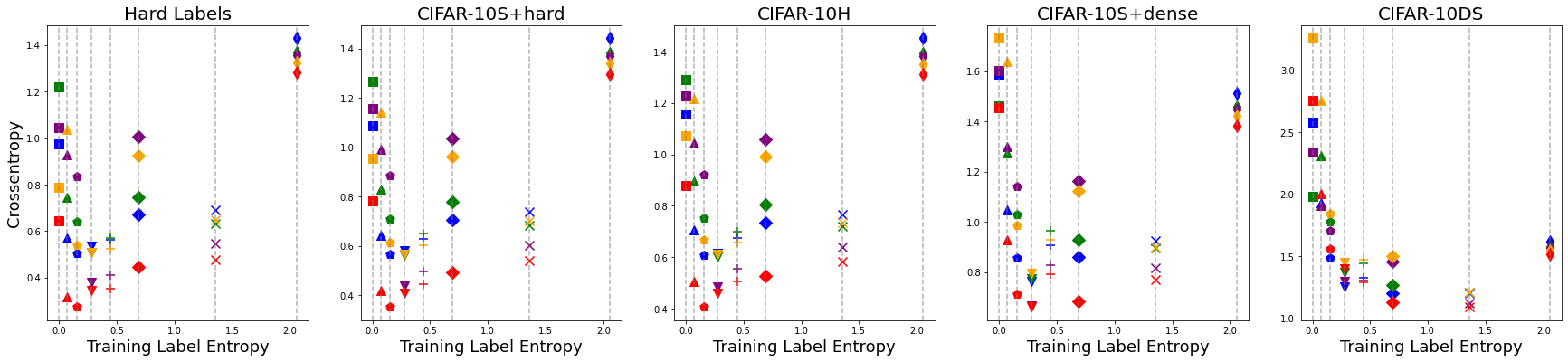}
    \includegraphics[width=0.9\textwidth]{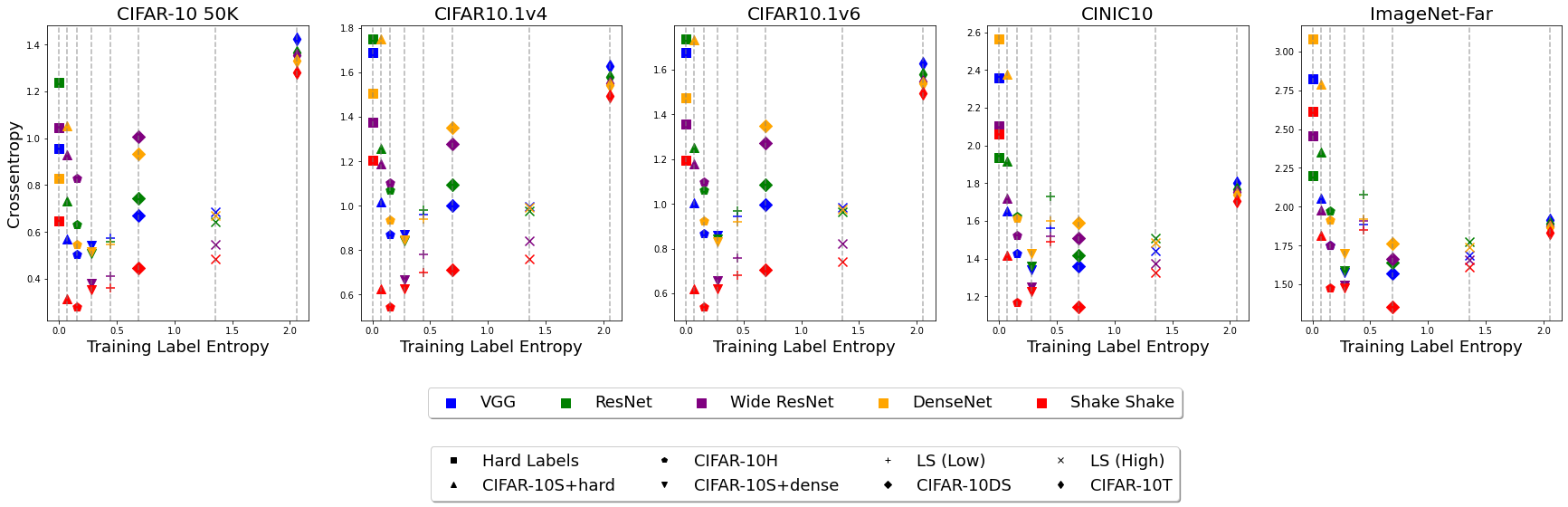}
    \vspace{-1mm}
    \caption{\textbf{Top:} Model performance on different label types at test time. \textbf{Bottom:} Generalization performance under increasing distributional shift. Each point represents the average score for a single model architecture (specified by color), trained on a particular label type (indicated via shape). Vertical lines represent points for a given label type.}
    \vspace{-3mm}
    \label{fig:genCheck}
\end{figure*}
\textbf{Models}: We train diverse image classifiers on each of our supervision sets. These models have distinct architectural features and reflect seminal developments in the progression of natural-image classification (see Supplement).

\textbf{Cross-Label Performance}:
We first assess how training classifiers on one set of soft labels impacts their validation performance when testing on other label types. Our primary measure of performance is crossentropy between the models' predictive distributions and soft labels. 
Consistent with the GNMDS experiments, we observe a U-shaped relationship between label entropy and model performance for nearly all soft-label types (Figure \ref{fig:genCheck}, top row). The one exception is when testing on \texttt{CIFAR-10DS} labels, where training on higher levels of softness is preferred. 
These results suggest that, for image classification, \textbf{\textit{sparse soft labels}}---of the kind collected in \cite{collins2022eliciting} or formed via averaging \citep{peterson2019human}---best capture the representational information expressed across different soft label sets. We also find that this relationship is preserved across different model architectures, with the \texttt{Shake Shake} model~\citep{gastaldi2017shake} performing best across all label sets.

\subsection{Zero-Shot Generalization}
To further test how well image classifiers extract representational information from soft labels, we assess zero-shot generalization performance on increasingly out-of-distribution image sets (Figure \ref{fig:genCheck}, bottom row; see Supplement). We find that for near-distribution datasets (\texttt{CIFAR-10 50K} \citep{krizhevsky2009learning}; \texttt{CIFAR10.1v6,v4} \citep{recht2018cifar}), classifier performance follows the U-shaped relationship described above. However, for far-distribution datasets (\texttt{CINIC} \citep{cinic}, \texttt{ImageNet-Far} \citep{peterson2019human}) a different pattern emerges. As found in \cite{peterson2019human}, \textbf{soft labels increasingly outperform hard labels as distribution shift increases}. 
The improved relative performance of the \texttt{CIFAR-10DS}-trained networks supports this -- although these labels are the most noisy, they also contain the richest supervision signal. It may be that more expressive classifiers are needed to fully capitalize on this richness \citep{battleday2020capturing, singh2020end}.

    

\subsection{Learning from small data} We empirically investigate our theoretical observation that softness is beneficial in the small data regime. Table \ref{tab:few-shot} presents model performance with 80 or 8 training examples per class. We find that soft labels outperform hard labels, and the same general pattern between softness and performance holds. 

\begin{table}[!b]
    \centering
    \vspace{-5mm}
    \caption{Small data regime results (crossentropy).} \label{tab:few-shot}
    \begin{tabular}{lcccc}
      \toprule 
      \bfseries Labels & \bfseries 80 l/c & \bfseries 8 l/c \\
      \midrule 
      \texttt{CIFAR-10} & 1.80 & 2.20\\
      \texttt{CIFAR-10S+hard} & 1.68 & 2.15\\
      \texttt{CIFAR-10H} & \textbf{1.67} & 2.15\\
      \texttt{CIFAR-10S+dense} & 1.69 & \textbf{2.12}\\
      \texttt{CIFAR-10LS (Low)} & 1.73 & 2.19\\
      \texttt{CIFAR-10DS} & 1.70 & \textbf{2.12}\\
      \texttt{CIFAR-10LS (High)} & 1.71 & 2.13\\
     \texttt{CIFAR-10T} & 2.10 & 2.25\\
      \bottomrule 
    \end{tabular}
\end{table}

\section{Discussion}
In this paper, we offer a principled set of theoretical and empirical findings aimed at helping researchers to determine which form of supervisory signal they ought to collect for the task at hand. We have provided theoretical grounding for how hidden representations can be recovered through supervised classification and have related the quality of these recovered representations to training parameters such as number of labels, classes, and dimensions. We found that while hard labels and soft labels provide comparable amounts of information in the many-examples-but-few-classes regime, soft labels become increasingly preferable when the number of classes increases or the number of labels decreases.  Our findings explain why, for example, pre-training a classifier on \texttt{ImageNet1K} (1,000 classes) or \texttt{ImageNet21k} (21,000 classes) using hard labels may lead to decent transfer learning performance~\citep{huh2016makes, ridnik2021imagenet} but pre-training with (a form of) soft labels may lead to even better transfer learning performance~\citep{Xie_2020_CVPR}. We support our theoretical contributions with empirical results on a suite of human-derived soft labels. We include a compilation of practical guidance around human soft label elicitation in the Supplement. We also note that our framework provides a general way to quantify the relative amount of information contained in each label by decomposing it into triplet queries (i.e., like a packet that contains a number of bits). This framework can be used with any representation learning setting where the goal is to efficiently recover hidden representations from an oracle (i.e., a person or another model) including classical supervised learning (which we focus on in this paper), knowledge distillation (getting a small student model to learn from a large teacher model), and contrastive learning (i.e., either via pairwise or triplet losses).

We note that, in our analysis, we made no assumptions about the data distribution in stimulus space, nor the function $f(x_i)=z_i$ that maps from stimulus space to hidden representations, but when training neural networks we often assume some level of stability or invariance (i.e., a small perturbation in pixel space does not lead to drastically different perception of the image). When satisfied, inductive biases like assumptions about stability or invariance allow learners to extract additional information from training examples, sometimes even in an unsupervised way when no labels are present. As a result, our analysis here can be considered as a sort of lower-bound on how much information about hidden representations a labeled training dataset can provide. We also examined each supervision signal in isolation, assuming that only labels of one type are collected. A promising future direction would be to analyze additional sources of information (inductive biases, other supervision signals, etc.) as well as the interactions between them; already, we see promising indications of mixing label types in the case of \texttt{CIFAR-10S+hard} and \texttt{CIFAR-10S+dense}. 

Notwithstanding these limitations, our analysis of hard labels, soft labels, and sparse labels that interpolate between them, already enables researchers to develop cost-benefit tradeoff curves in order to optimize the cost of labeling their datasets for supervised learning---and support the development of data-efficient, generalizable ML systems. 

\begin{acknowledgements} 


    This work was made possible by support from the NOMIS Foundation and the Office of Naval Research (N00014-18-1-2873) to TLG and an NSERC fellowship
(567554-2022) to IS. KMC gratefully acknowledges support from the Marshall Commission and the Cambridge Trust. UB  acknowledges  support  from  DeepMind  and  the  Leverhulme Trust  via  the  Leverhulme  Centre  for  the  Future  of  Intelligence  (CFI),  and  from  the  Mozilla  Foundation. AW  acknowledges  support  from  a  Turing  AI  Fellowship  under grant  EP/V025279/1,  The  Alan  Turing  Institute,  and  the Leverhulme Trust via CFI.
\end{acknowledgements}

\bibliography{sucholutsky_617}
\onecolumn 
\vspace{10mm}


\appendix

\section{Practical guidance on human soft label elicitation}

Our framework provides users with a way to quantify the relative amount of information contained in each type of label so that they can optimize which labels to collect for their dataset. Specifically, the findings from our theory, simulations, and experimental results are that the relative informativeness of labels depends on three factors associated with the dataset (the number of labeled examples, the number of classes, and the latent dimensionality) and two factors associated with labels (error rate and sparsity). Out of these, most factors are not under the user’s control but the primary factor that users can control (in supervised learning settings) is label sparsity. Our guidance to users is thus the following:
\begin{itemize}
    \item Use our framework to estimate the relative informativeness of the label types you are considering collecting. The key parameter to optimize is label sparsity so compute the informativeness of soft labels with different levels of sparsity. 
    \item Pick out promising label types and run a small pilot study collecting each label type for a small set of objects. Compute error rates and per-label costs for each label type. 
    \item Update the relative informativeness estimates based on error rates and calculate the cost-benefit tradeoff for each label type. Pick the type with the most favorable tradeoff.

\end{itemize}
For users who want a simpler procedure, we offer the following rule-of-thumb guidance:
\textit{
Generally, softer labels are preferable in smaller data regimes (e.g. one-shot and less-than-one-shot learning) while harder labels are preferable in big data regimes (i.e. many-shot learning).
}

\section{Label optimization simulations}
\begin{figure*}[htb!]
    \centering
    \includegraphics[width=0.24\textwidth]{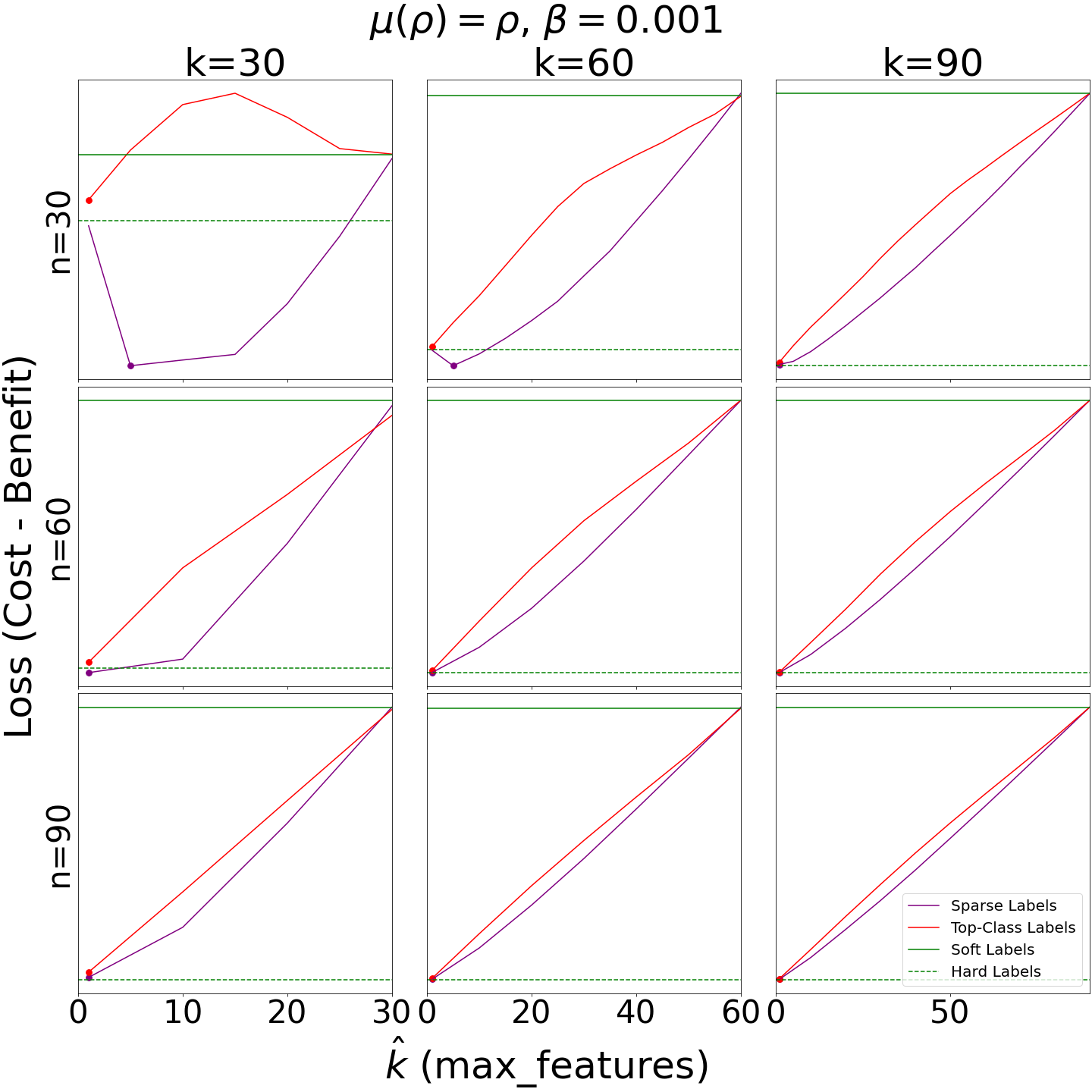}
    \includegraphics[width=0.24\textwidth]{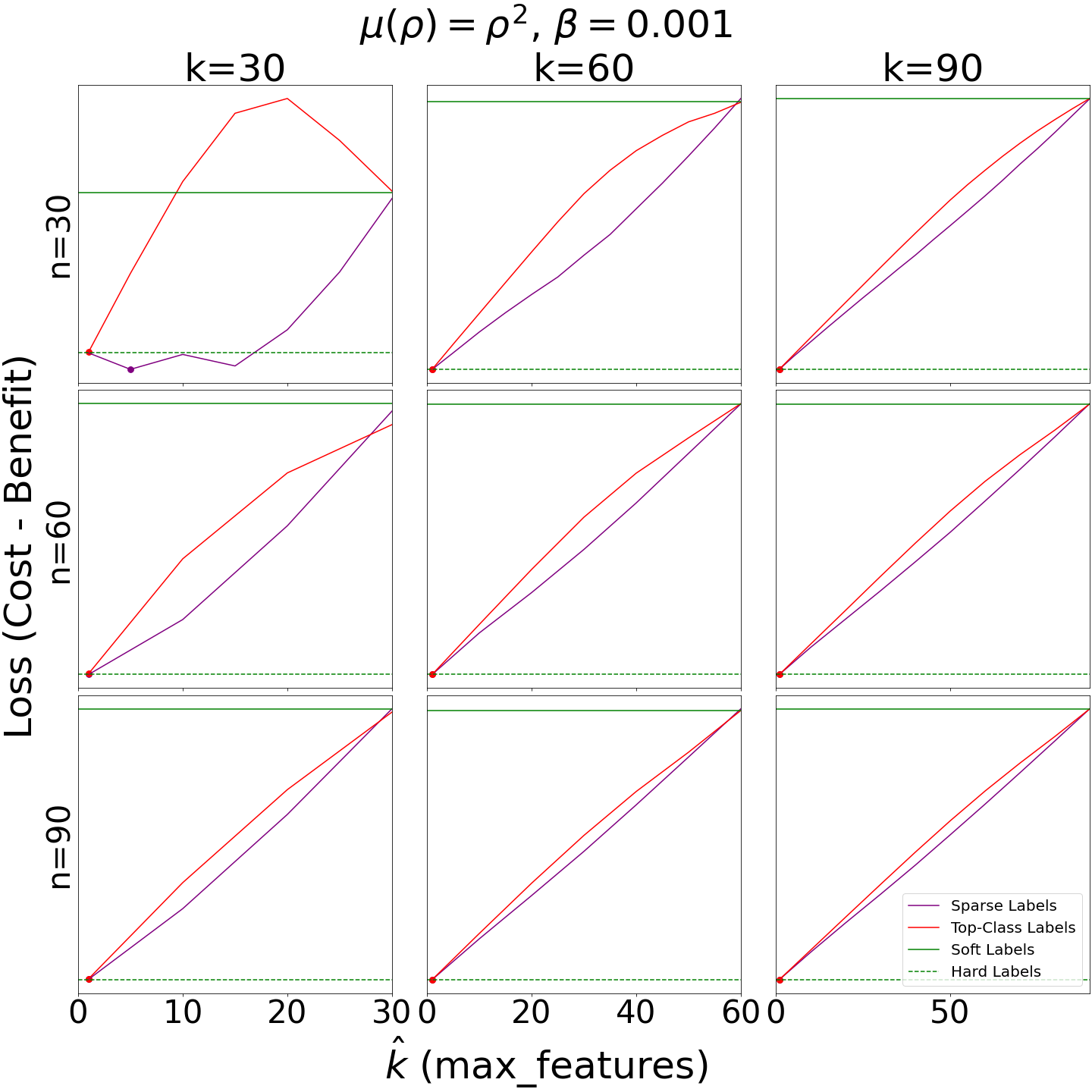}
    \includegraphics[width=0.24\textwidth]{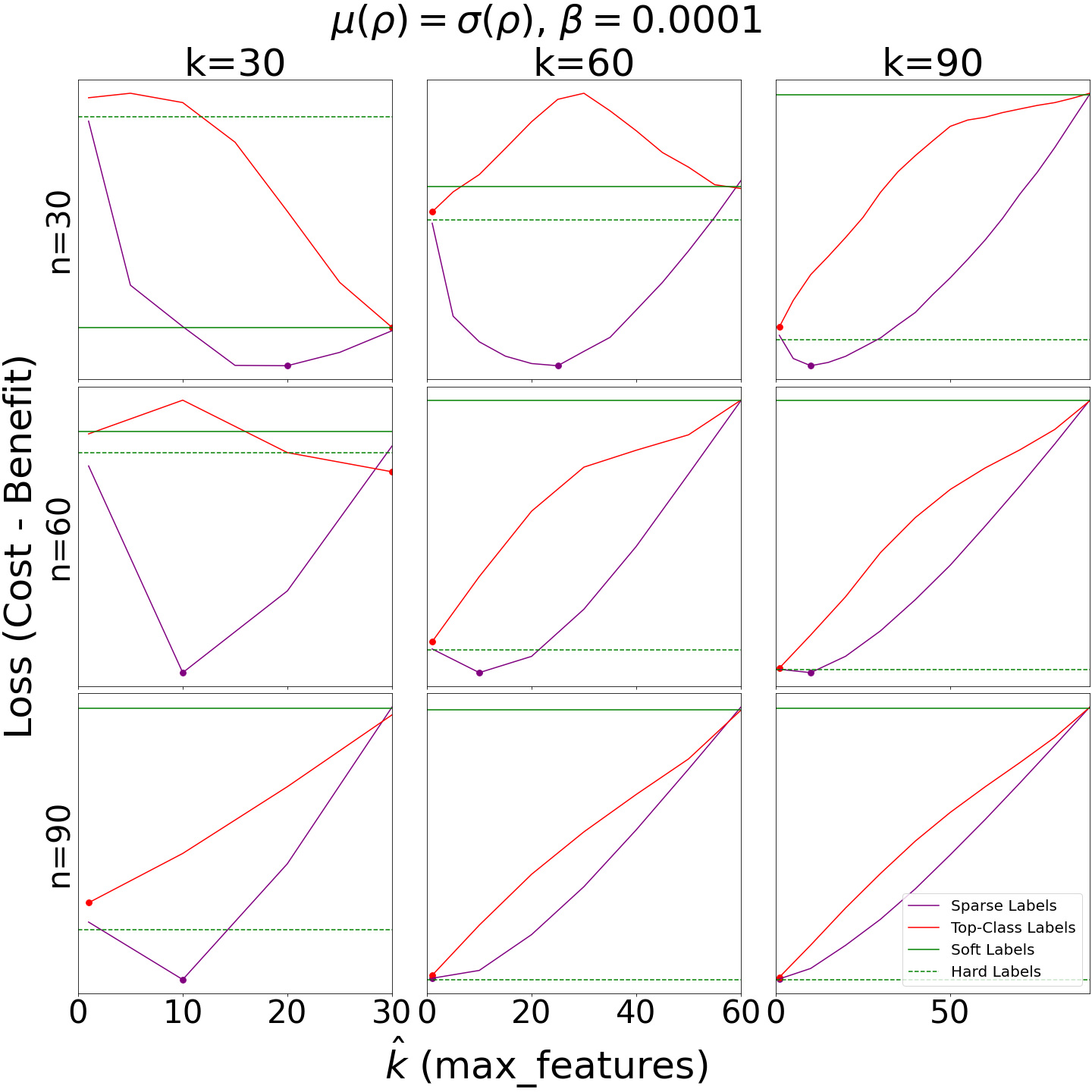}
    \includegraphics[width=0.24\textwidth]{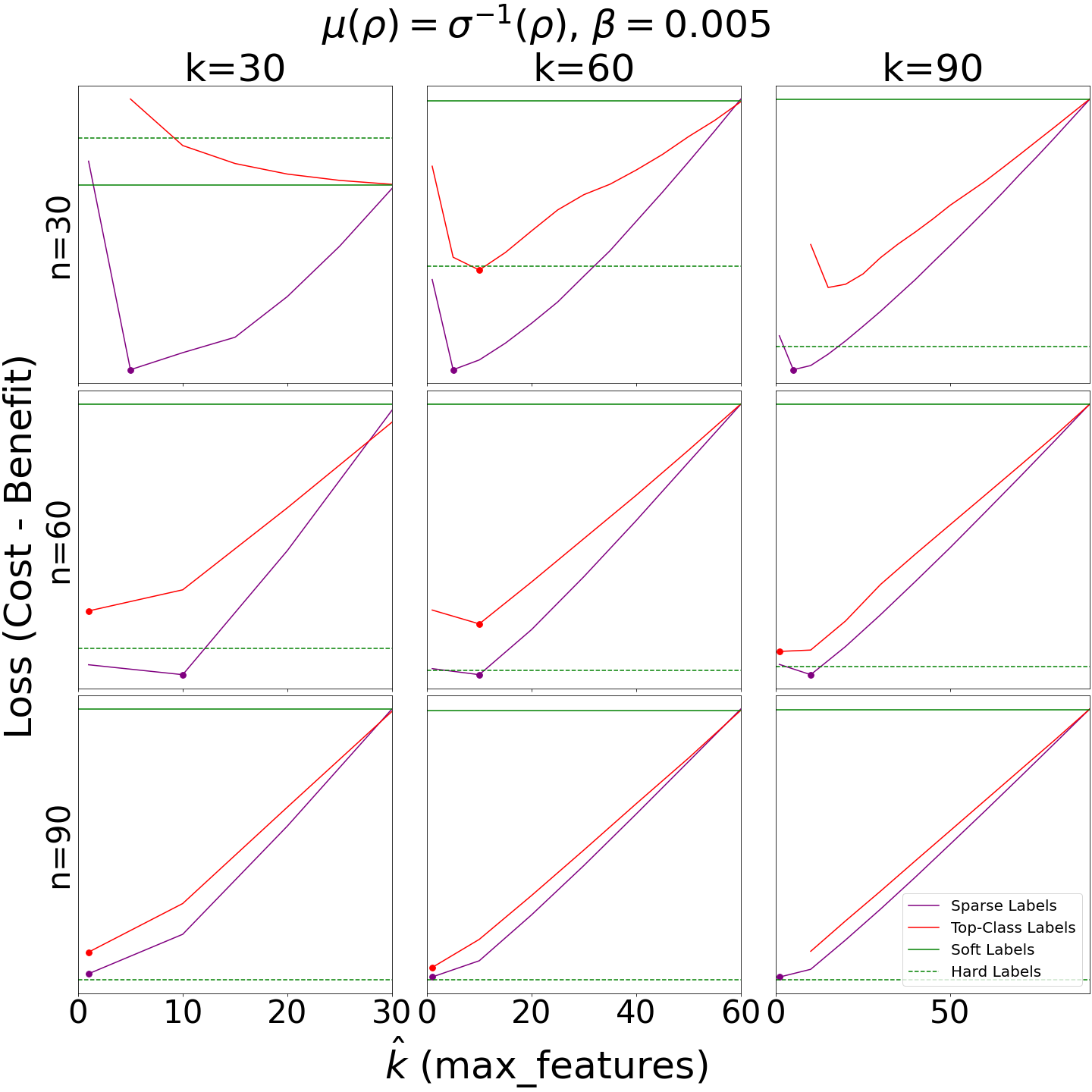}
    \includegraphics[width=0.24\textwidth]{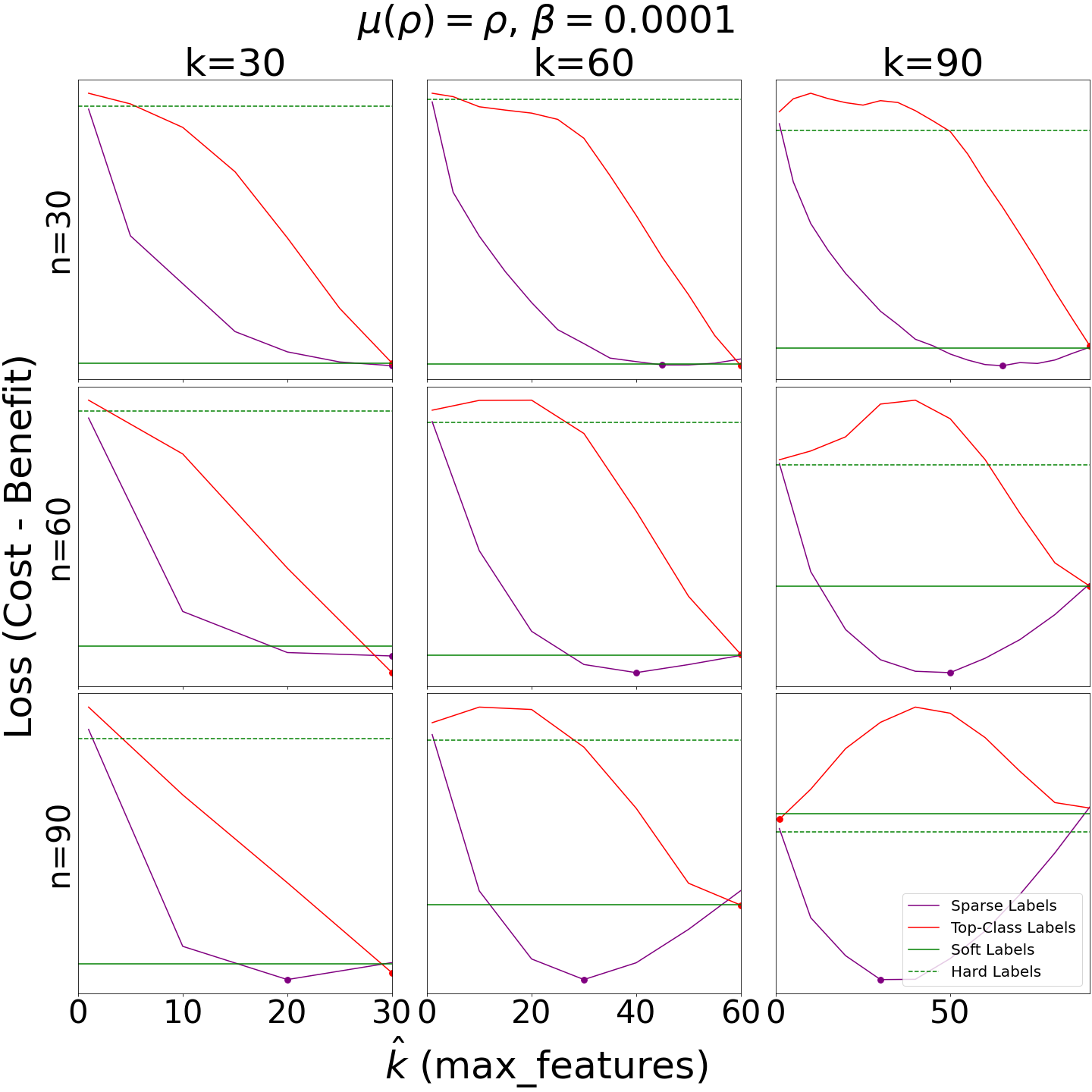}
    \includegraphics[width=0.24\textwidth]{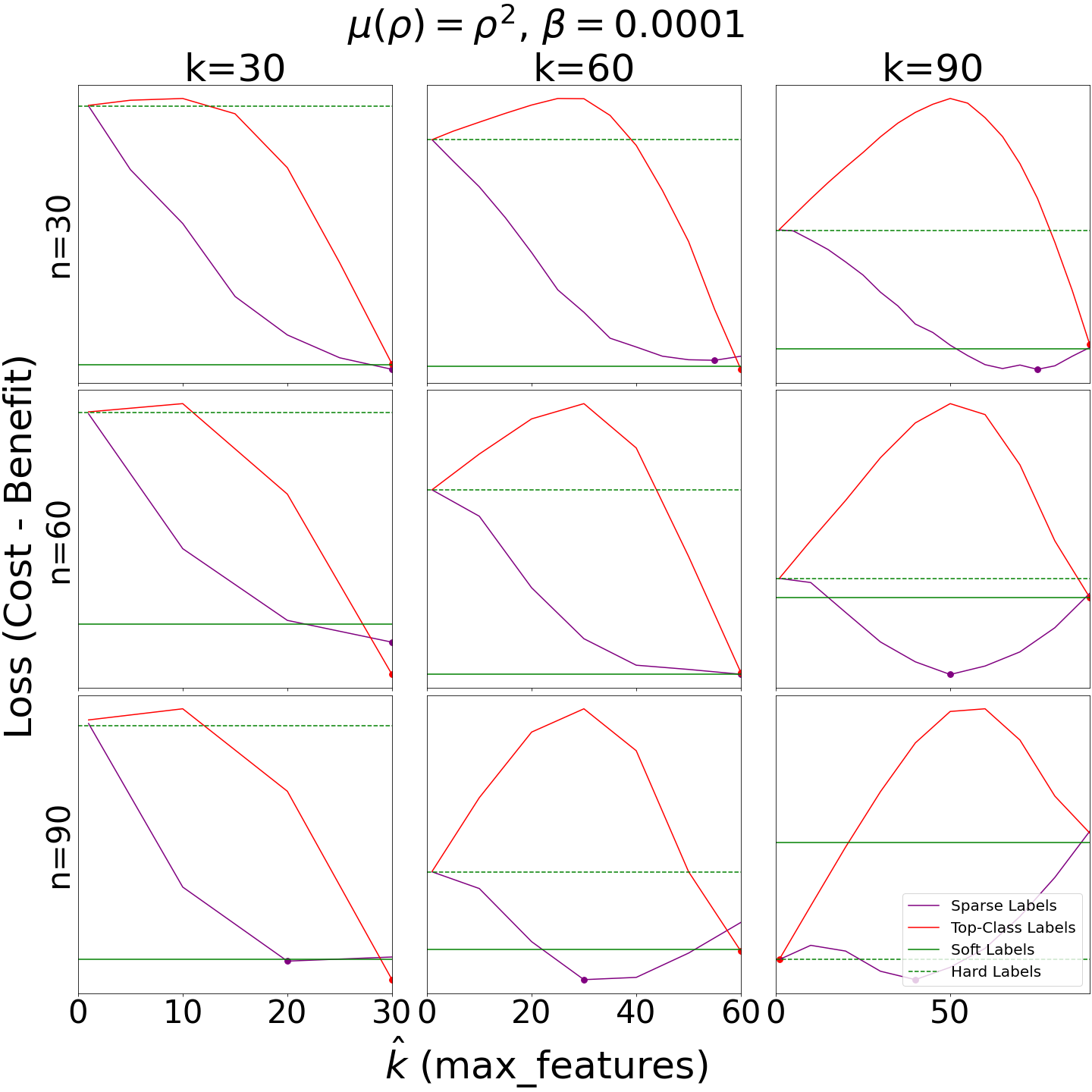}
    \includegraphics[width=0.24\textwidth]{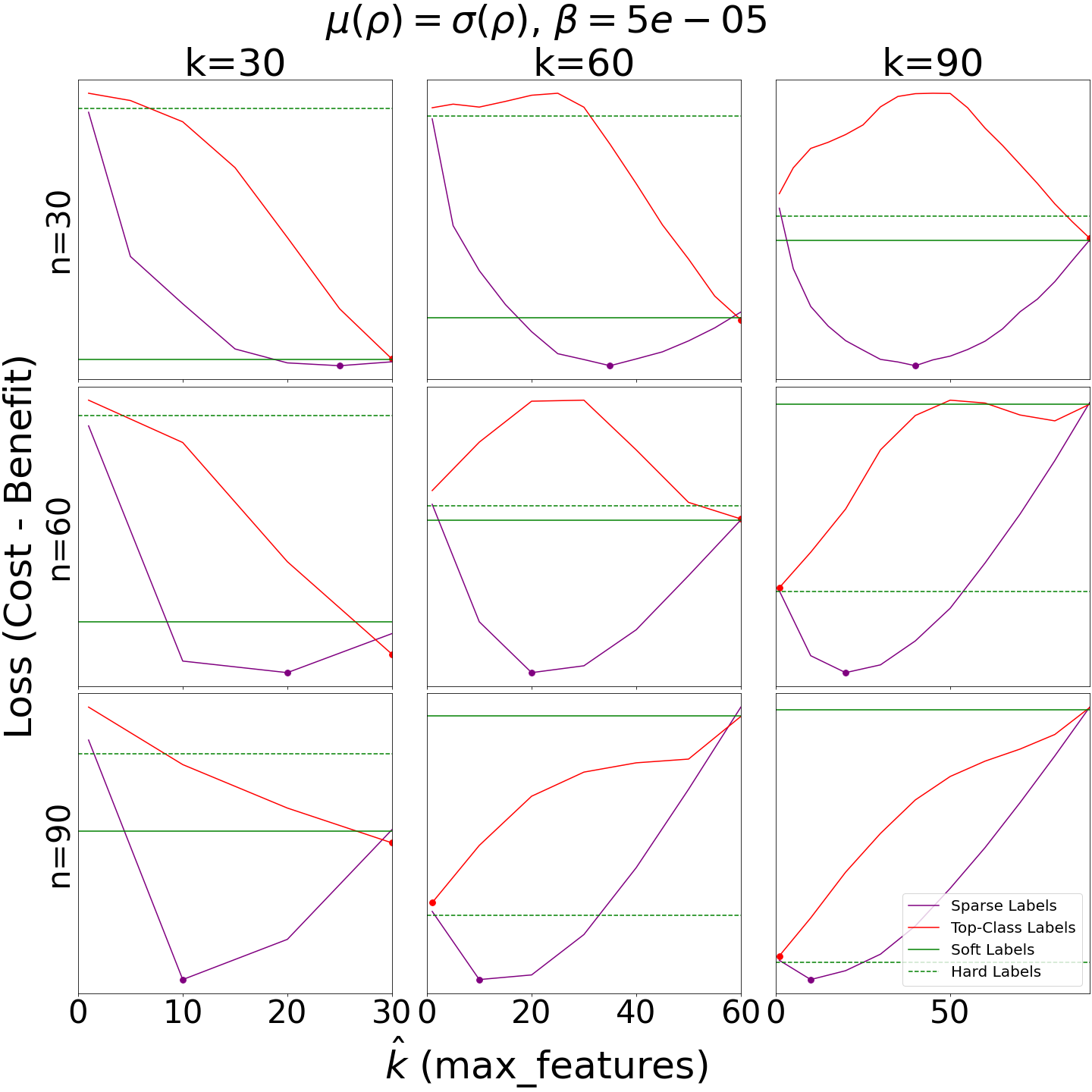}
    \includegraphics[width=0.24\textwidth]{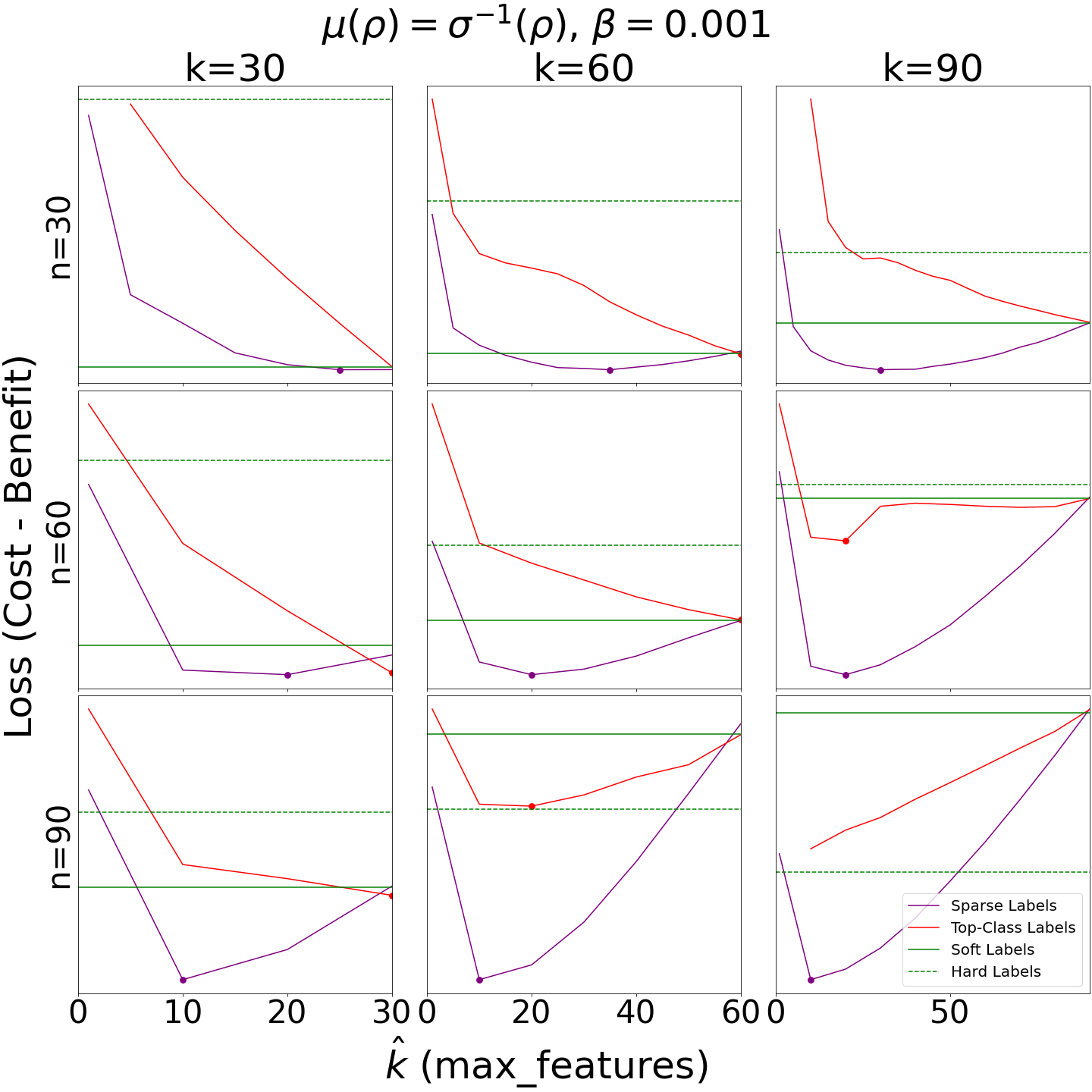}
    \caption{Loss curves for soft labels (solid green), sparse labels (purple), top-class labels (red), and hard labels (dashed green) based on subjective utility function ($u(\rho)$), cost weighting parameter ($\beta$), sparsity ($\hat k$), number of points ($n$), and number of classes ($k$). Global minima for sparse labels and top-class labels are marked with a point.}
    \label{fig:fig4}
\end{figure*}

See Figure~\ref{fig:fig4}.

\section{Elicited-Human vs Model-Predicted Entropy}
\begin{figure*}[htb!]
    \centering
    \includegraphics[width=0.18\textwidth]{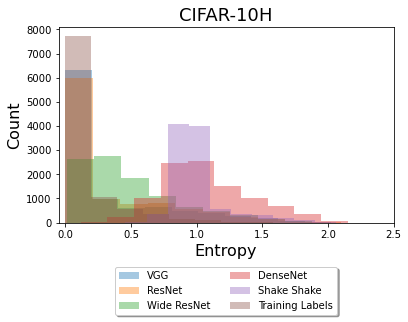}
    \includegraphics[width=0.18\textwidth]{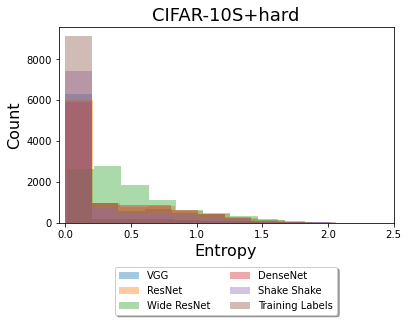}
    \includegraphics[width=0.18\textwidth]{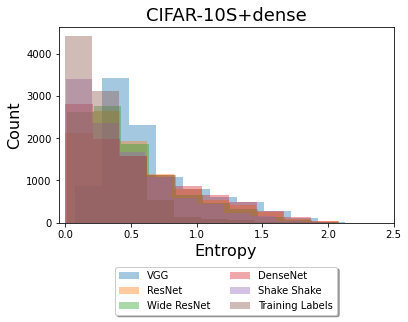}
    \includegraphics[width=0.18\textwidth]{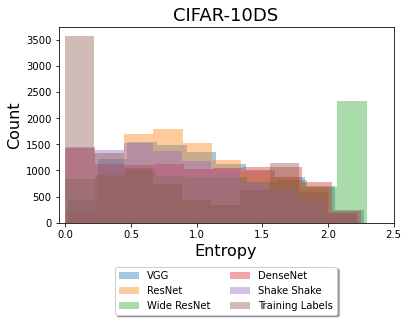}
    \includegraphics[width=0.18\textwidth]{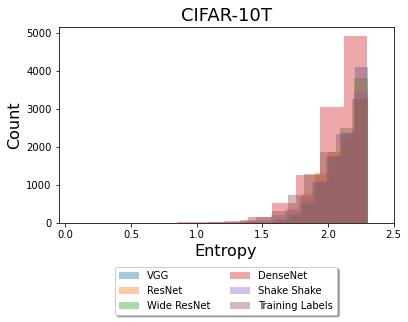}
    \includegraphics[width=0.18\textwidth]{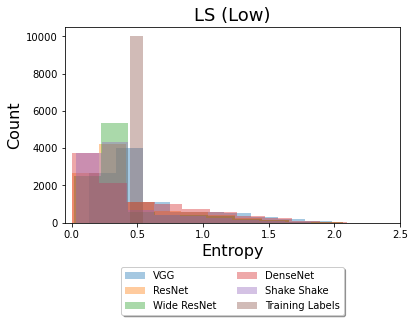}
    \includegraphics[width=0.18\textwidth]{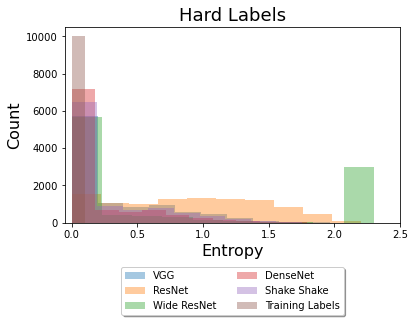}
    \includegraphics[width=0.18\textwidth]{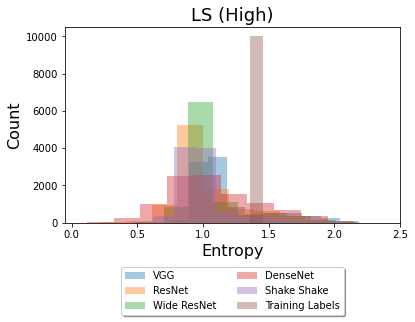}
    \caption{Comparing the entropy of the models' predictions against the entropy of the labels used to train them. The training label type is listed as the title for the respective histogram.}
    \label{fig:modeltraininglabelent}
\end{figure*}

We investigate the hypothesis that the labels which confer the best downstream performance may strike a natural resonance with the models they are used as supervision signals for. In Figure~\ref{fig:modeltraininglabelent}, we compare the entropy of the training labels against the entropy of the trained models' predicted distributions. In other words, we compare the probability distributions produced by each model, to the probability distributions that the model was trained on. We find a remarkable alignment between the entropy of models' predictions trained on the \texttt{CIFAR-10S} varieties. Future work could investigate the links between the inductive biases of models and the labels best suited for training specific architectures. 


\section{Additional Details on Human Soft Labels} 
\subsection{Collecting \texttt{CIFAR-10DS} and Similarity Judgments}
Soft labels for \texttt{CIFAR-10DS}, as well as similarity judgments, were collected on Amazon Mechanical Turk (AMT). The recruitment and experimental pipelines were automated using the PsyNet framework for online experiment design \cite{harrison2020gibbs}. Prior to participation in the studies, participants provided informed consent in accordance with an Institutional Review Board (IRB), and were paid at a rate of \$12 per hour. In addition, participants were required to have successfully completed at least 2000 tasks on AMT.

To collect \texttt{CIFAR-10DS}, participants observed individual images and were given a set of sliders (10, one for each category) ranging from 0 to 1 and were asked to move the sliders in accordance with how well they thought each category matched a given image, with 0 being ``not at all matching'' and 1 being ``completely matching''. We aimed for about 10 multi-ratings per image and each participant completed 50 such multi-ratings.

As for similarity judgments, participants were presented with pairs of unlabeled images and were required to rate their similarity on a 7-point Likert scale ranging from 0 (``completely dissimilar'') to 6 (``completely similar''). Here we aimed for 5 judgments per pair of images and each participant completely an average of 80 such judgments.

\subsection{In-Filling \texttt{CIFAR-10S} Labels}
The \texttt{CIFAR-10S} labels collected in \cite{collins2022eliciting} included only 1,000 of the full 10,000 \texttt{CIFAR-10} test set. Note, however, that these 1,000 examples were already enriched to be those that are naturally more confusing -- so it can be considered a sensible sampling of what -10S labels \textit{may} look like more generally. However, for adequate comparison against the other label types, we needed to choose a labeling method to label the remaining 9,000. We elected two variants: 1) using hard labels, or 2) simulating \texttt{CIFAR-10S} labels via sparsified version of \texttt{CIFAR-10DS}. The former represents a real-world cost efficient scenario; we could imagine a researcher only having the budget to annotate a subset of a dataset with soft labels. The second case is designed to mimic what the labels may have been like had we elicited \texttt{CIFAR-10S} over the full set. Taking only the scalar values for the top two highest sliders from \texttt{CIFAR-10DS} offered a nice entropy- and conceptual-match (entropy of 0.69 for \texttt{CIFAR-10S} to 0.75 for the adjusted \texttt{-10DS} labels). Future work could explore automated measures to extend label conversions (e.g., learning a mapping from \texttt{CIFAR-10DS} to simulated \texttt{CIFAR-10S} labels). We note that the \texttt{CIFAR-10S} labels used in this work are the T2 Clamp varieties, with a redistribution factor of $10\%$ following \citeauthor{collins2022eliciting}.

\textbf{CIFAR-10T} The \texttt{CIFAR-10T} labels are a novel set of labels we crowdsourced, comprising over 350,000 typicality ratings for each image under the ground truth category (about 35  judgements per image). 1759 unique participants were recruited on Amazon Mechanical Turk, and presented with a sequence of 200 randomly sampled \texttt{CIFAR-10} test set images, upsampled to 160x160 pixels (see \cite{peterson2019human, battleday2020capturing}. Participants were given the category of each image, and asked to rate how
typical it was of the category on a sliding scale
of ``Not at all typical" to ``Extremely typical". We interpret an image's typicality as the probability of the ground truth class, and spread the remaining probability mass over the 9 remaining labels---a smoothed version of a \textit{sparse} soft label with K=1).

\subsection{Additional Similarity Judgment Studies}
\begin{figure}[htb!]
    \centering
    \includegraphics[width=0.8\textwidth]{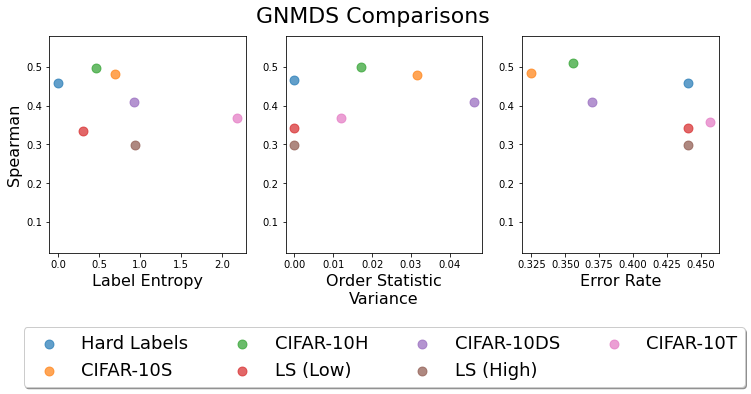}
    \caption{Correlation between ground-truth similarity judgments and the cosine similarity of image representations for different model architectures.}
    \label{fig:gnmdsHumanOther}
\end{figure}

We extend the GNMDS analyses in the main text by examining the similarity structure of image representations extracted from the penultimate layer of each network. For each image and network, we derive an abstract vector representation by storing the unit activations of the last layer during classification. Then, for each image we compute the pairwise cosine similarity between the representations derived from our classifiers. We correlate these to the ground truth similarity ratings, and present the results in Figure \ref{fig:gnmdsHumanOther}. The images used for these analyses are discussed below, and displayed in Figures \ref{fig:entropy_images} and \ref{fig:relentropy_images}.

\section{Additional Computational Experiment Details and Observations} 

\subsection{Models}
We use ten fold cross-validation to partition the images of the \texttt{CIFAR-10} test subset into train and validation sets for each set of soft labels. We train a number of models using stochastic gradient descent over a range of learning rates and seeds, and use the best performing seed for all subsequent analyses (Table \ref{tab:models}). We use ten fold cross-validation to partition the images of the \texttt{CIFAR-10} test subset into train and validation sets for each set of soft labels. 

\begin{table}[!h]
    \centering
    \caption{Image Classifiers.} \label{tab:models}
    \begin{tabular}{rccc}
      \toprule 
      \bfseries Model & \bfseries Key Features & \bfseries  Parameters & \bfseries Citation \\
      \midrule 
      VGG & very deep connections & 14{,}728{,}266 & \cite{simonyan2014very}\\
      ResNet & residual connections &  & \cite{he2016deep}\\
      WRN & wide residual connections & 36{,}479{,}194 & \cite{zagoruyko2016wide}\\
      DenseNet & dense connections & 769{,}162 & \cite{huang2017densely}\\
      Shake shake & shake shake regularization & 11{,}709{,}514& \cite{gastaldi2017shake}\\
      \bottomrule 
    \end{tabular}
\end{table}

\subsection{Datasets}
 In order of increasing distributional shift, \texttt{CIFAR-10 50K} is the \textit{training} subset of \texttt{CIFAR10} (50{,}000 images; \cite{krizhevsky2009learning}, \texttt{CIFAR10.1v6,v4} are two near-sample datasets constructed from the same TinyImages classes \cite{torralba200880} as \texttt{CIFAR-10} (2{,}000 images each; \cite{recht2018cifar}), our subset of \texttt{CINIC10} contains rescaled images from ImageNet using the \texttt{CIFAR-10} classes (210{,}000 images; \cite{cinic}), and \texttt{ImageNet-Far} contains a label-coarsened version of rescaled \texttt{ImageNet} images such that the CIFAR classes now contain a more diverse range of examples (for example, now ``deer'' contains ``ibex'' and ``gazelle''; 63{,}895 images; \cite{peterson2019human, cinic}). 




\subsection{Softness, Task Accuracy, and Information Content}

\begin{figure*}[t!]
    \centering
    \includegraphics[width=0.9\textwidth]{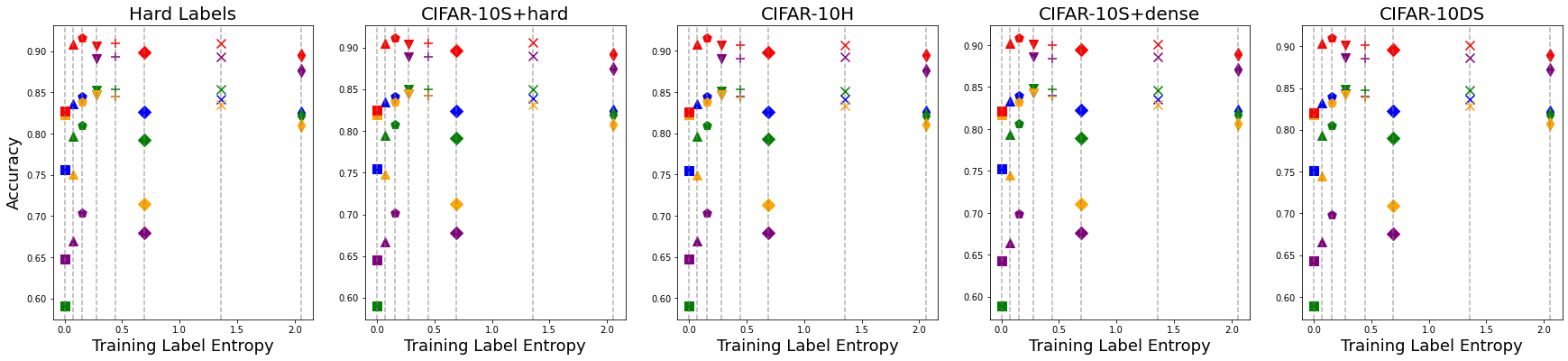}
    \includegraphics[width=0.9\textwidth]{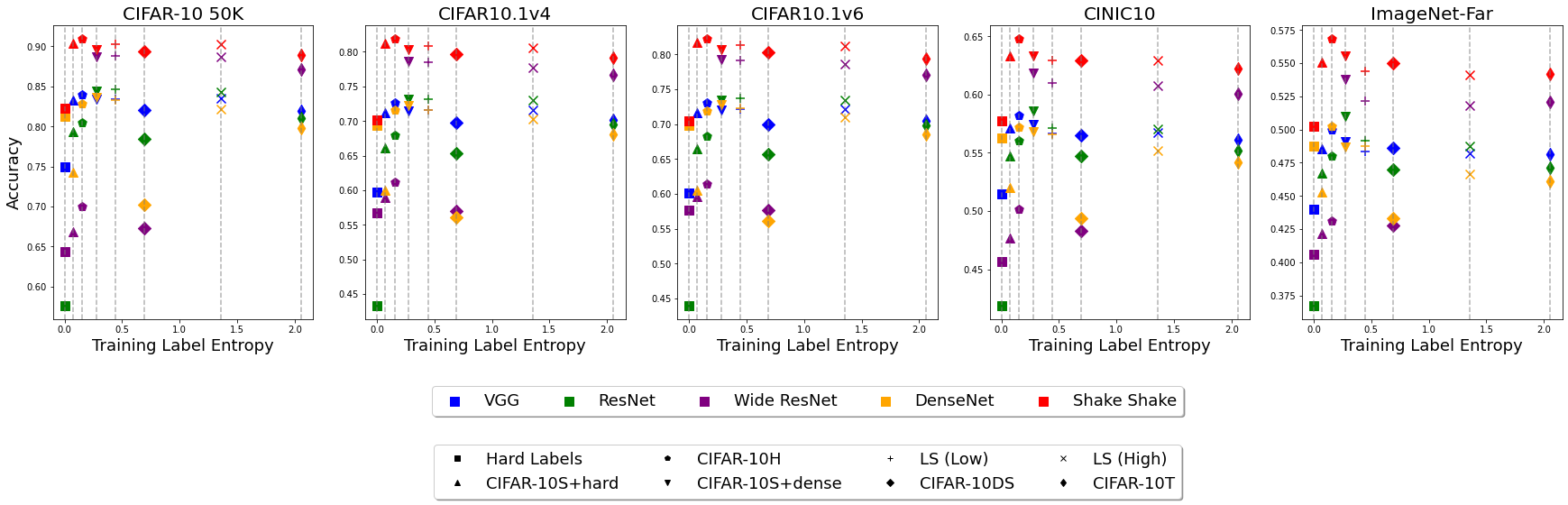}
    \vspace{-1mm}
    \caption{Cross-label (top) and generalization results (bottom), scored by top-1 accuracy against the respective labels. }
    \vspace{-3mm}
    \label{fig:full_score_acc}
\end{figure*}

In the main text, we depicted the relationship between label softness and task performance using crossentropy (CE) as our principal metric. We focus on CE as it better captures the fidelity of the models’ predictive distributions. This is particularly important when we evaluate the model on held-out soft labels; CE offers more information about model performance than just top-1 accuracy. However, we include top-1 accuracy in Figure \ref{fig:full_score_acc} for completeness.

\begin{figure*}[t!]
    \centering
    \includegraphics[width=0.9\textwidth]{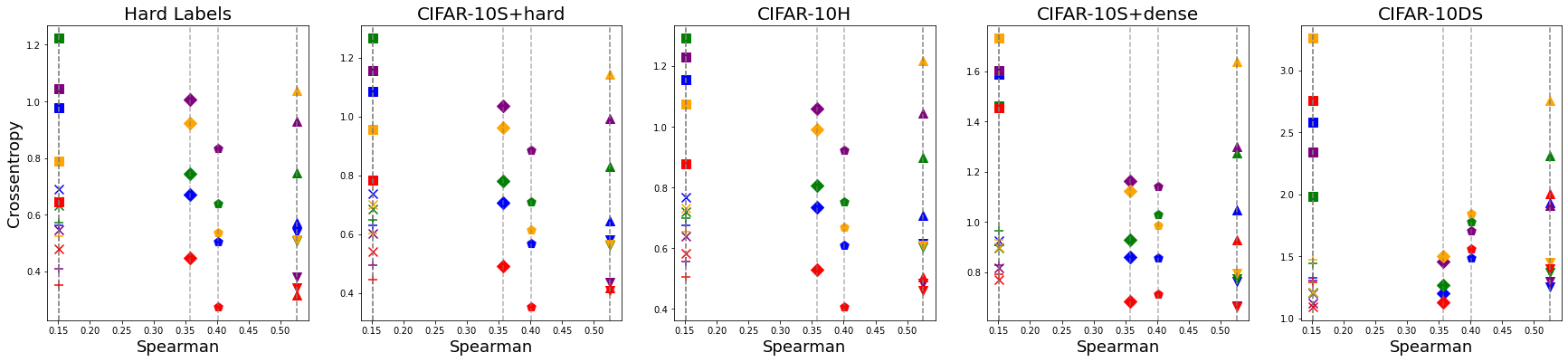}
    \includegraphics[width=0.9\textwidth]{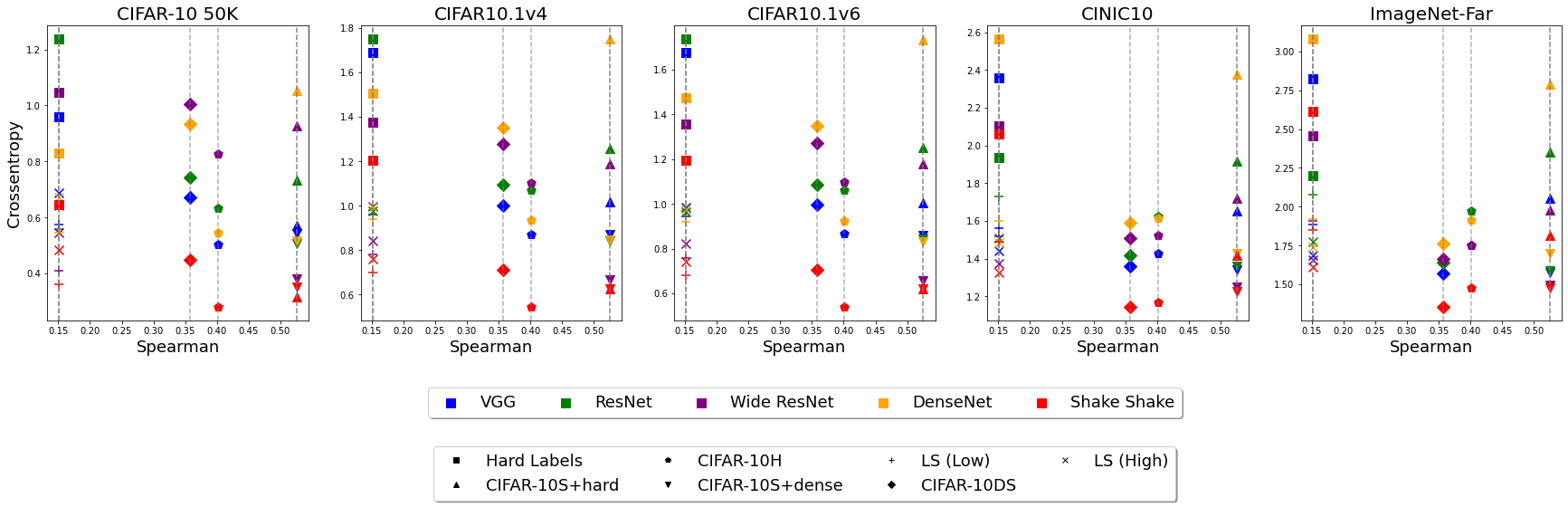}
    \vspace{-1mm}
    \caption{\textbf{Top:} Model performance on different label types at test time, as a function of information content of the labels. Information content is approximated by the Spearman rank correlation with similarity judgments. \textbf{Bottom:} Generalization performance under increasing distributional shift, as a function of training label information content.}
    \vspace{-3mm}
    \label{fig:infoContentPerf}
\end{figure*}

We also depict performance in Figure \ref{fig:infoContentPerf} as function of the information content of the labels. Here, we use the Spearman rank correlation coefficient between the \texttt{CIFAR-10} GNMDS and elicited similarity judgments as a proxy for information content. Note, here, \texttt{CIFAR-10S+hard} and \texttt{CIFAR-10S+dense} have the same score, as the similarity judgments are collected over the 1000 shared original \texttt{CIFAR-10S} examples.

\subsection{Effective dimensionality of soft labels}
Our theory and simulations address the number of features available for representation learning but did not discuss the nature of these features—i.e., whether they are essential or superfluous. Estimating the effective dimensionality of a dataset is tied to the nature of the computational task required. For classification, there is a range of methods for estimating this (e.g., \citep{doi:10.1137/1116025,jha2023extracting}).

\subsection{Varying Label Softness}

We further investigate how the amount of softness we elicit from humans when constructing our supervision signals impacts downstream performance: a selective ablation for increasing levels of sparsity. In our real-world soft label experiments, we in-fill missing \texttt{CIFAR-10S} labels by simulating if \texttt{CIFAR-10DS} labels had only provided uncertainty over $\hat{k}=2$ labels. As we have softness over all $k=10$, we can simulate varying $\hat{k}$. We train additional models in-filling with $\hat{k} = 3$ and $4$, respectively. We find that the extra softness does not add substantial value over $\hat{k}=2$ when evaluating on near-domain generalization (i.e. CIFAR10-50k, .1v4, and .1v6), but does appear to have a positive effect when evaluating on further out-of-domain generalization (e.g., CINIC10 and ImageNet-Far). 


\begin{table}[]
    \centering
    \caption{Crossentropy (lower is better) as a function of varying the classes we permit human uncertainty specification over.}
    \label{tab:varyK}
    \begin{tabular}{@{}llllll@{}}
    \toprule
    Classes per Label & CIFAR10-50k & CIFAR10.1v4 & CIFAR10.1v6 & CINIC10 & ImageNet-Far \\ \midrule
    k = 2             & \textbf{0.46}        & \textbf{0.77}      & \textbf{0.76}       & 1.32     & 1.57         \\
    k = 3             & 0.48        & \textbf{0.77}        & \textbf{0.76}        & 1.30     & 1.51         \\
    k = 4             & 0.49        & \textbf{0.77}        & 0.77        & \textbf{1.26}     & \textbf{1.47}         \\
    k = 10            & 0.76        & 1.09        & 1.08        & 1.40     & 1.60         \\ \bottomrule
    \end{tabular}
    
\end{table}

\subsection{Representative images and model predictions}
\begin{figure}[h!]
    \centering
    \includegraphics[width=\textwidth]{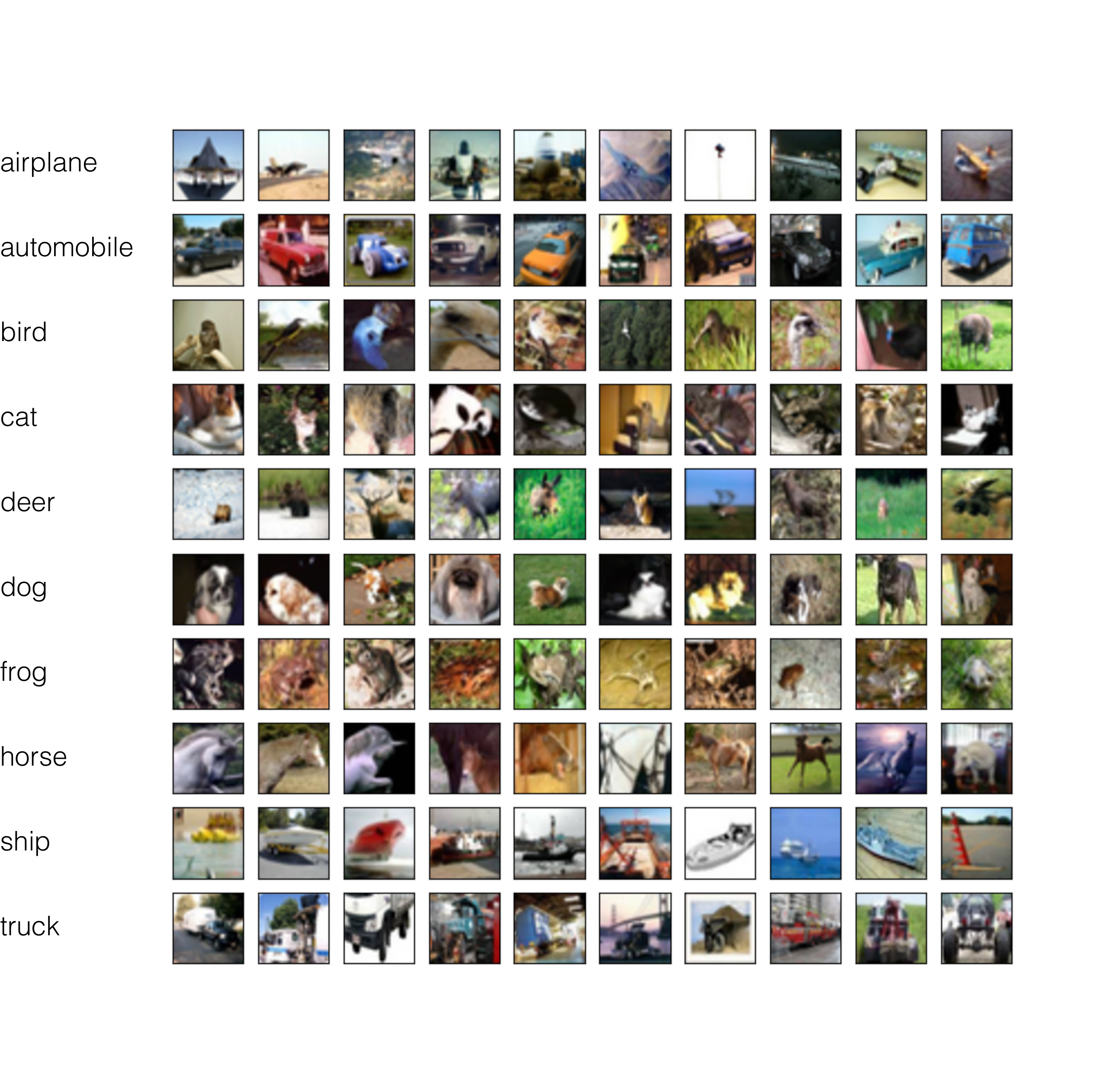}
    \caption{100 images from the \texttt{CIFAR-10} testing subset. These were chosen to include images that were likely to have high label entropy.}
    \label{fig:entropy_images}
\end{figure}

\begin{figure}[h!]
    \centering
    \includegraphics[width=\textwidth]{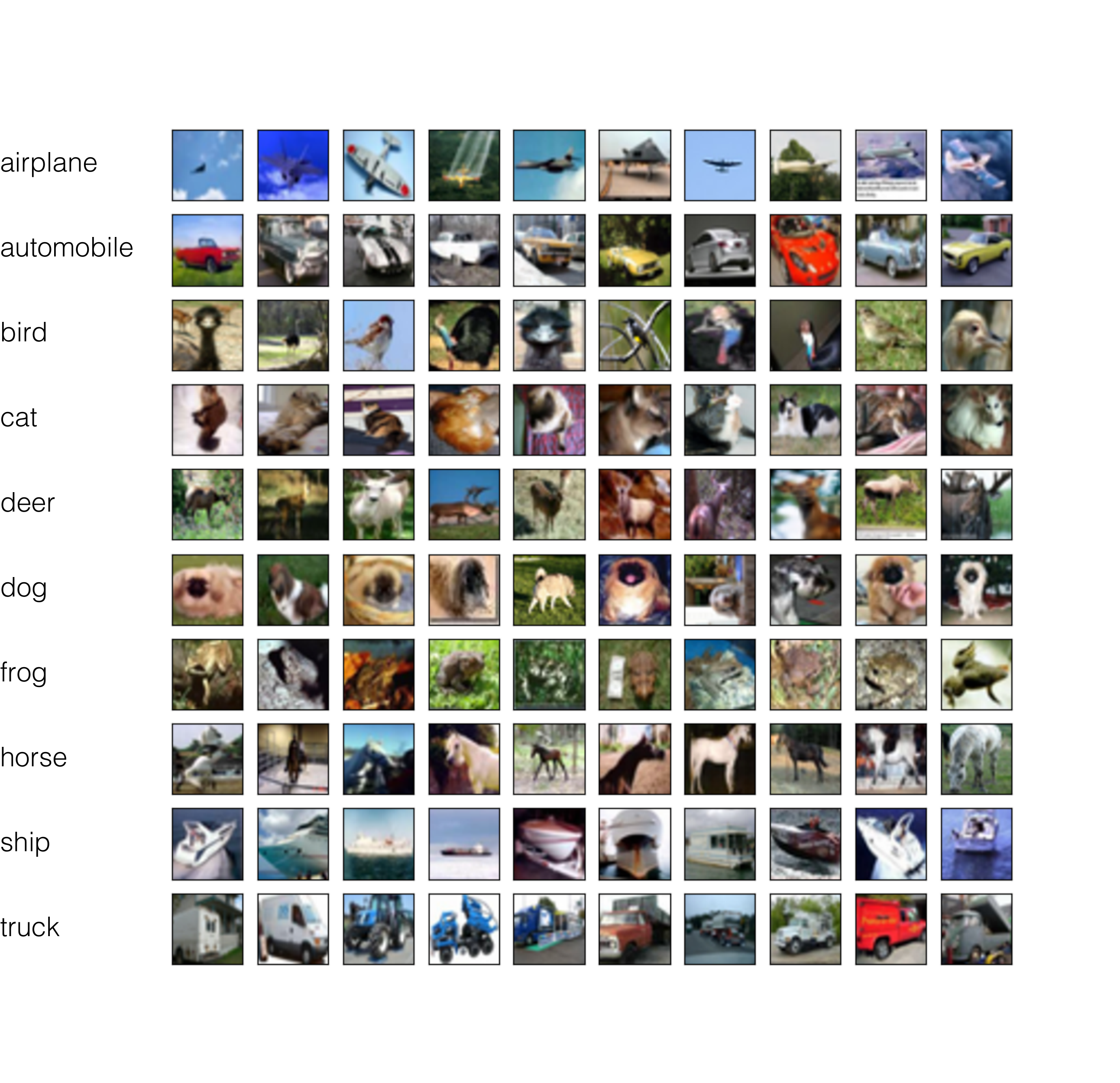}
    \caption{100 images from the \texttt{CIFAR-10} testing subset. These were chosen to include images that were likely to cause model disagreement.}
    \label{fig:relentropy_images}
\end{figure}

In Figures \ref{fig:entropy_images} and \ref{fig:relentropy_images} we present the images used as the basis of the similarity experiments (see above for details on label and similarity judgment collection). These images were chosen \textit{using our trained classification models} to include images that were likely to have high label entropy (Figure \ref{fig:entropy_images}), and images where model predictions diverged (Figure \ref{fig:relentropy_images}). The models making the predictions had not been trained on these images (i.e., the predictions were based on the held-out cross-validation folds). 

\begin{figure}[h!]
    \centering
    \includegraphics[width=0.9\textwidth]{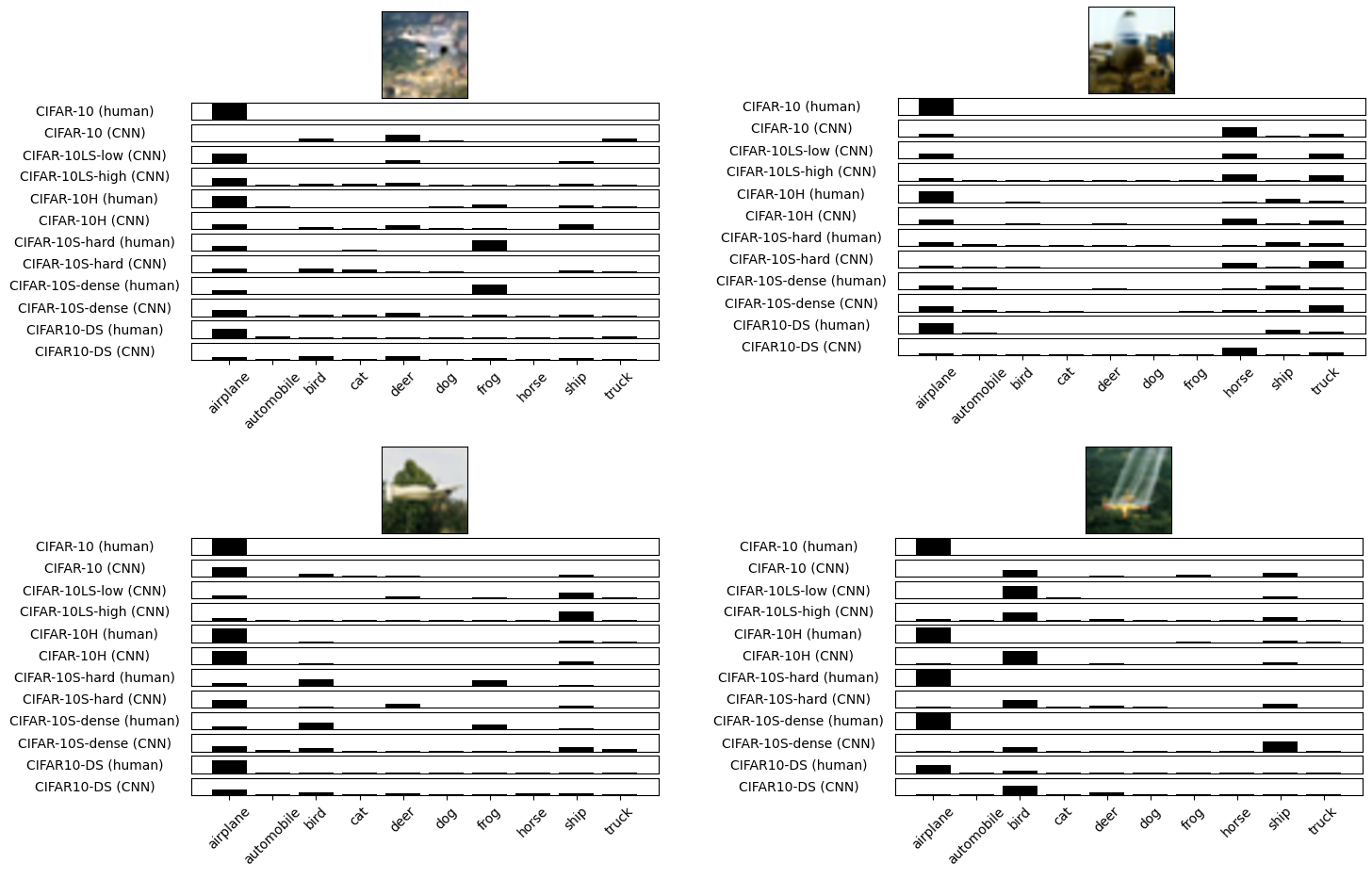}
    \caption{Representative ambiguous plane images. Top row: model agreement on high entropy image. Bottom row: maximal model disagreement.}
    \label{fig:plane}
\end{figure}

\begin{figure}[h!]
    \centering
    \includegraphics[width=0.9\textwidth]{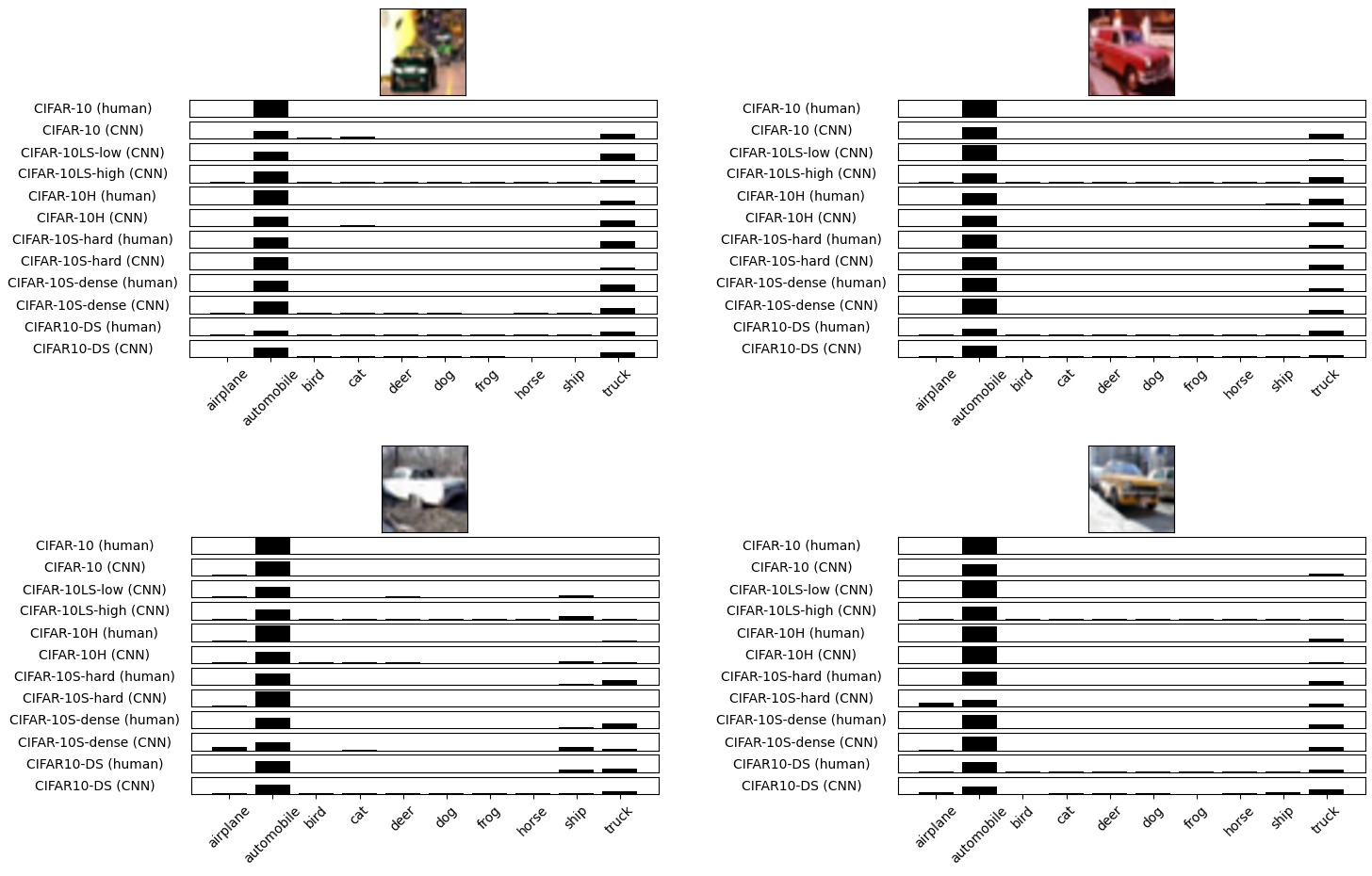}
    \caption{Representative ambiguous automobile images. Top row: model agreement on high entropy image. Bottom row: maximal model disagreement.}
    \label{fig:auto}
\end{figure}

\begin{figure}[h!]
    \centering
    \includegraphics[width=0.9\textwidth]{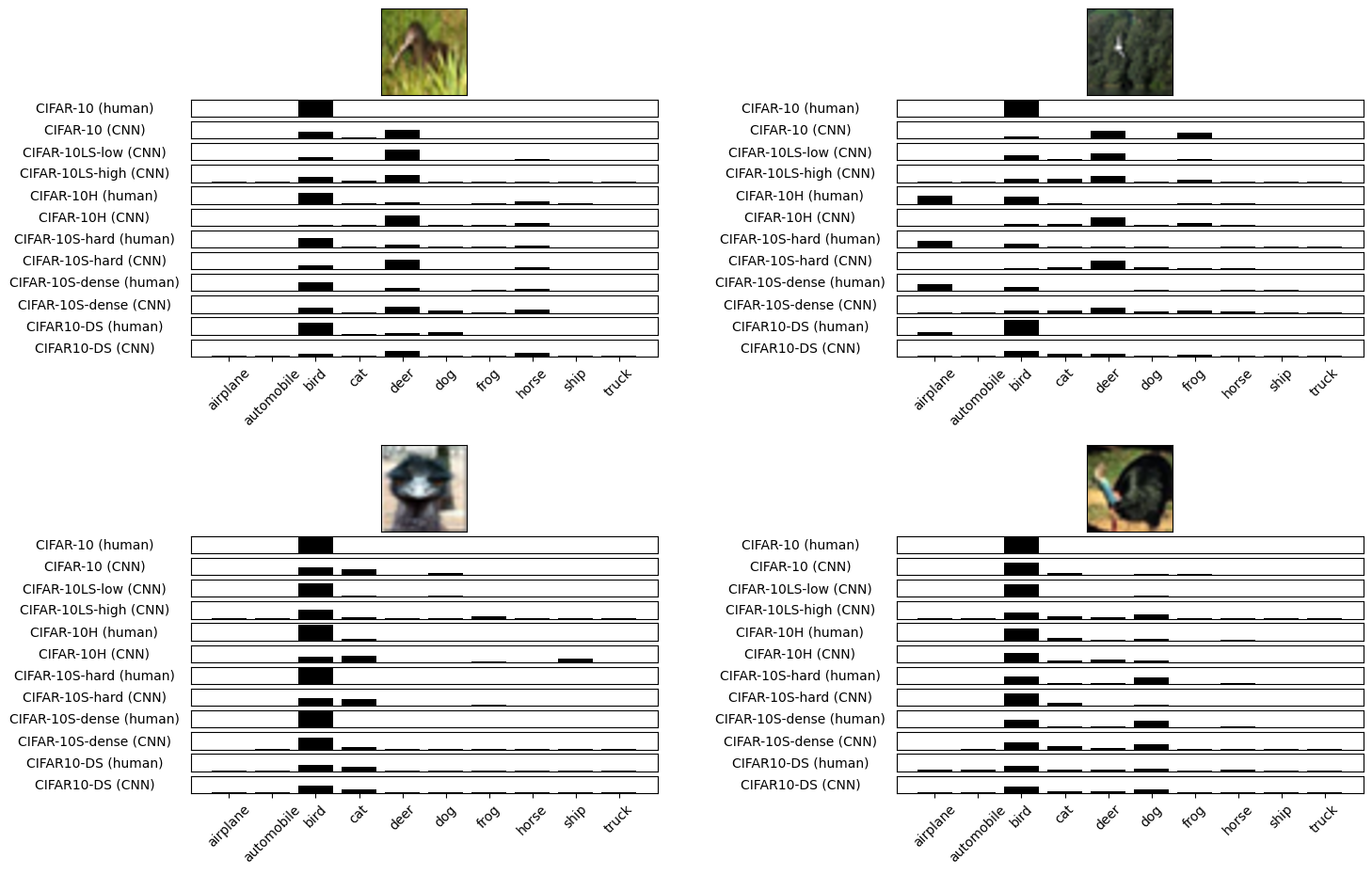}
    \caption{Representative ambiguous bird images. Top row: model agreement on high entropy image. Bottom row: maximal model disagreement.}
    \label{fig:bird}
\end{figure}

\begin{figure}[h!]
    \centering
    \includegraphics[width=0.9\textwidth]{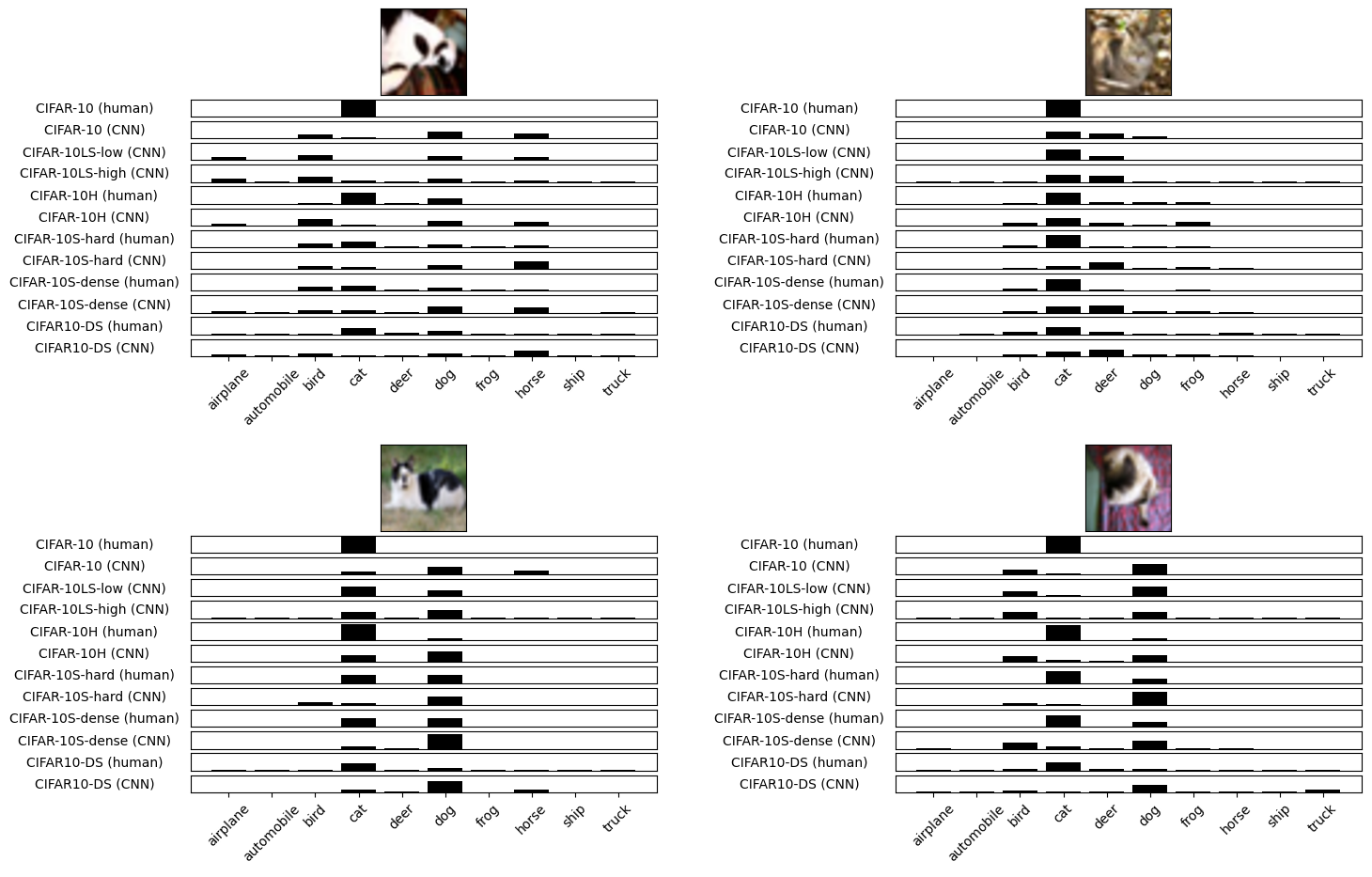}
    \caption{Representative ambiguous cat images. Top row: model agreement on high entropy image. Bottom row: maximal model disagreement.}
    \label{fig:cat}
\end{figure}

\begin{figure}[h!]
    \centering
    \includegraphics[width=0.9\textwidth]{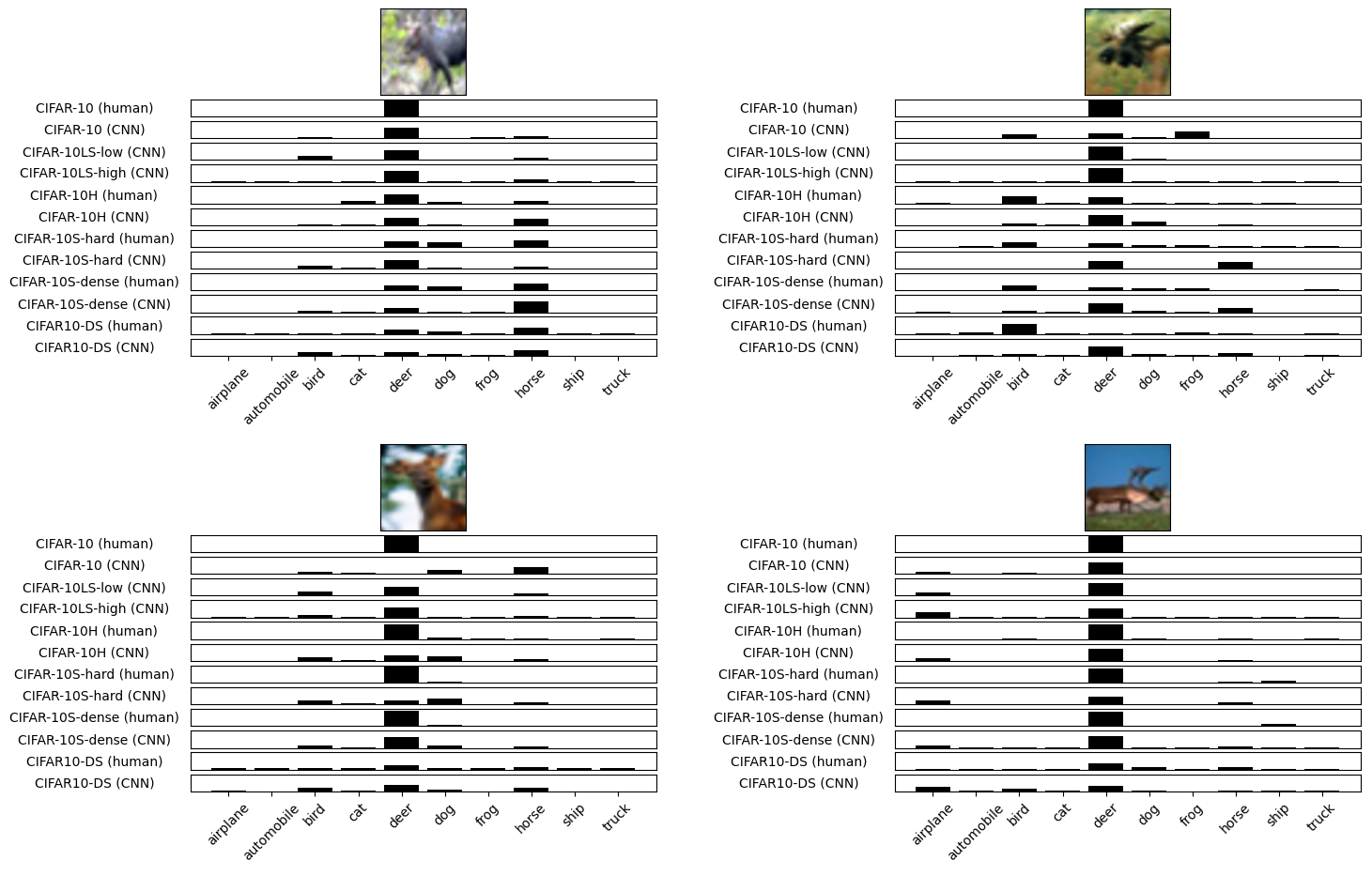}
    \caption{Representative ambiguous deer images. Top row: model agreement on high entropy image. Bottom row: maximal model disagreement.}
    \label{fig:deer}
\end{figure}

\begin{figure}[h!]
    \centering
    \includegraphics[width=0.9\textwidth]{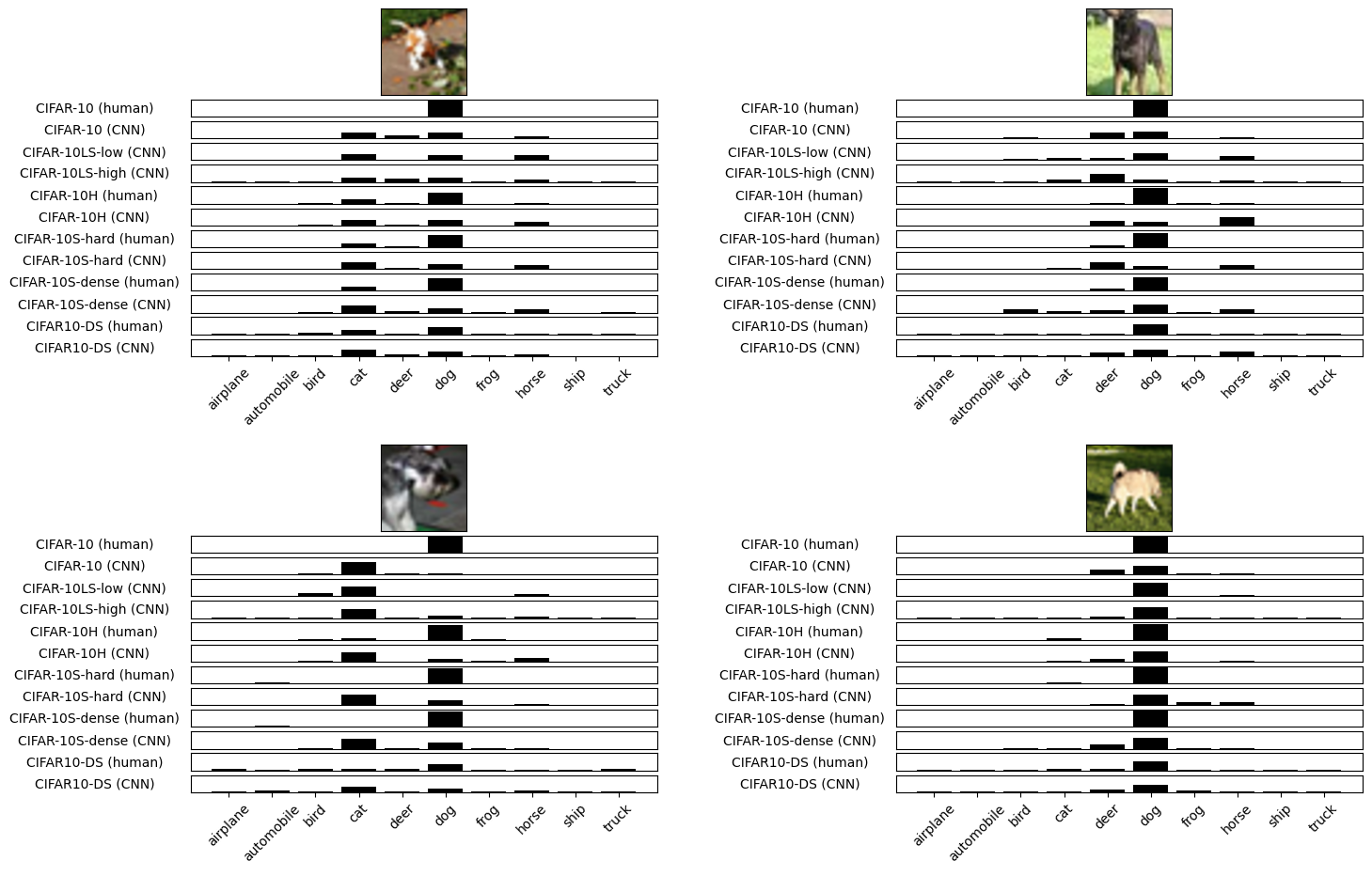}
    \caption{Representative ambiguous dog images. Top row: model agreement on high entropy image. Bottom row: maximal model disagreement.}
    \label{fig:dog}
\end{figure}

\begin{figure}[h!]
    \centering
    \includegraphics[width=0.9\textwidth]{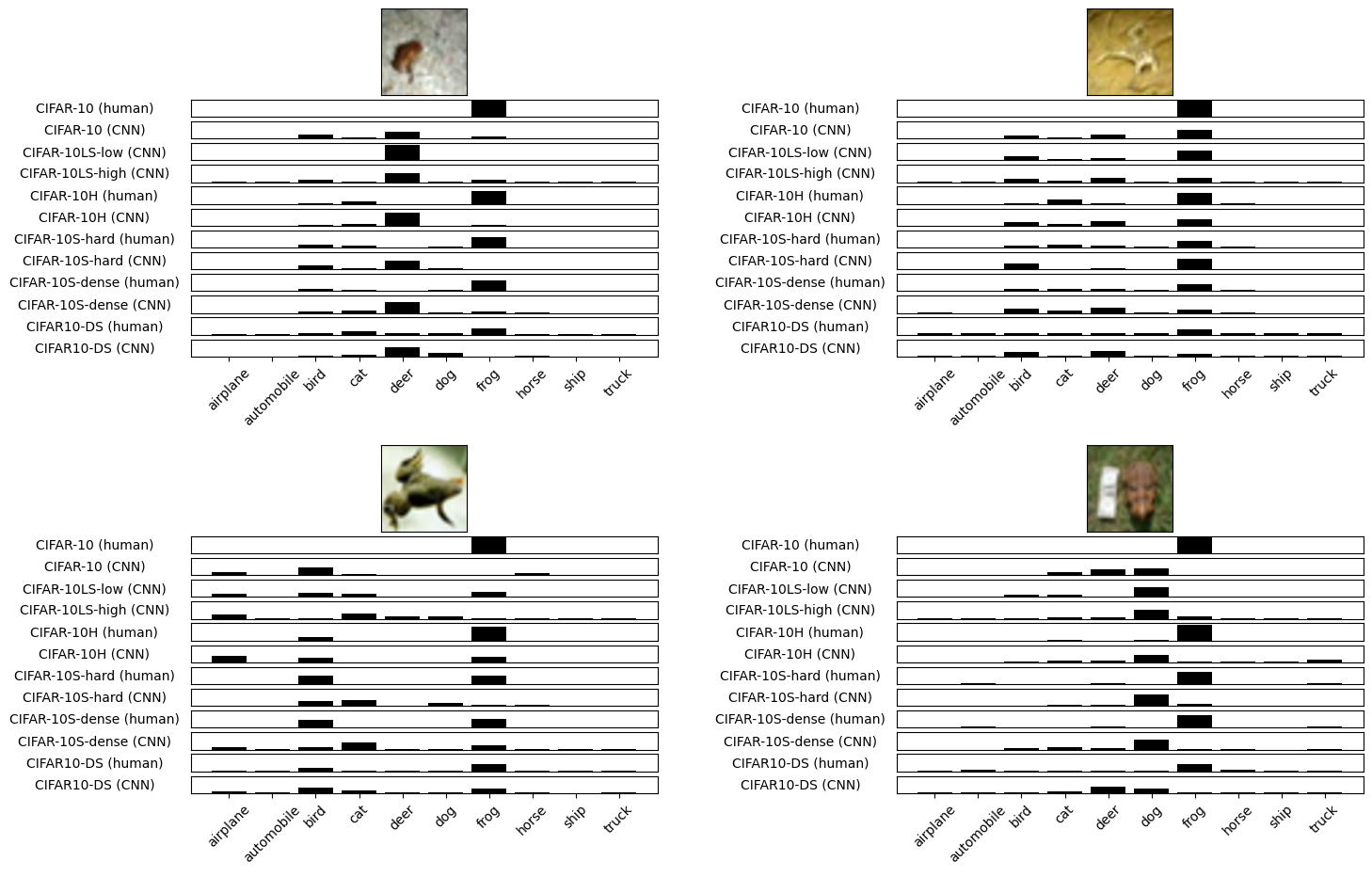}
    \caption{Representative ambiguous frog images. Top row: model agreement on high entropy image. Bottom row: maximal model disagreement.}
    \label{fig:frog}
\end{figure}

\begin{figure}[h!]
    \centering
    \includegraphics[width=0.9\textwidth]{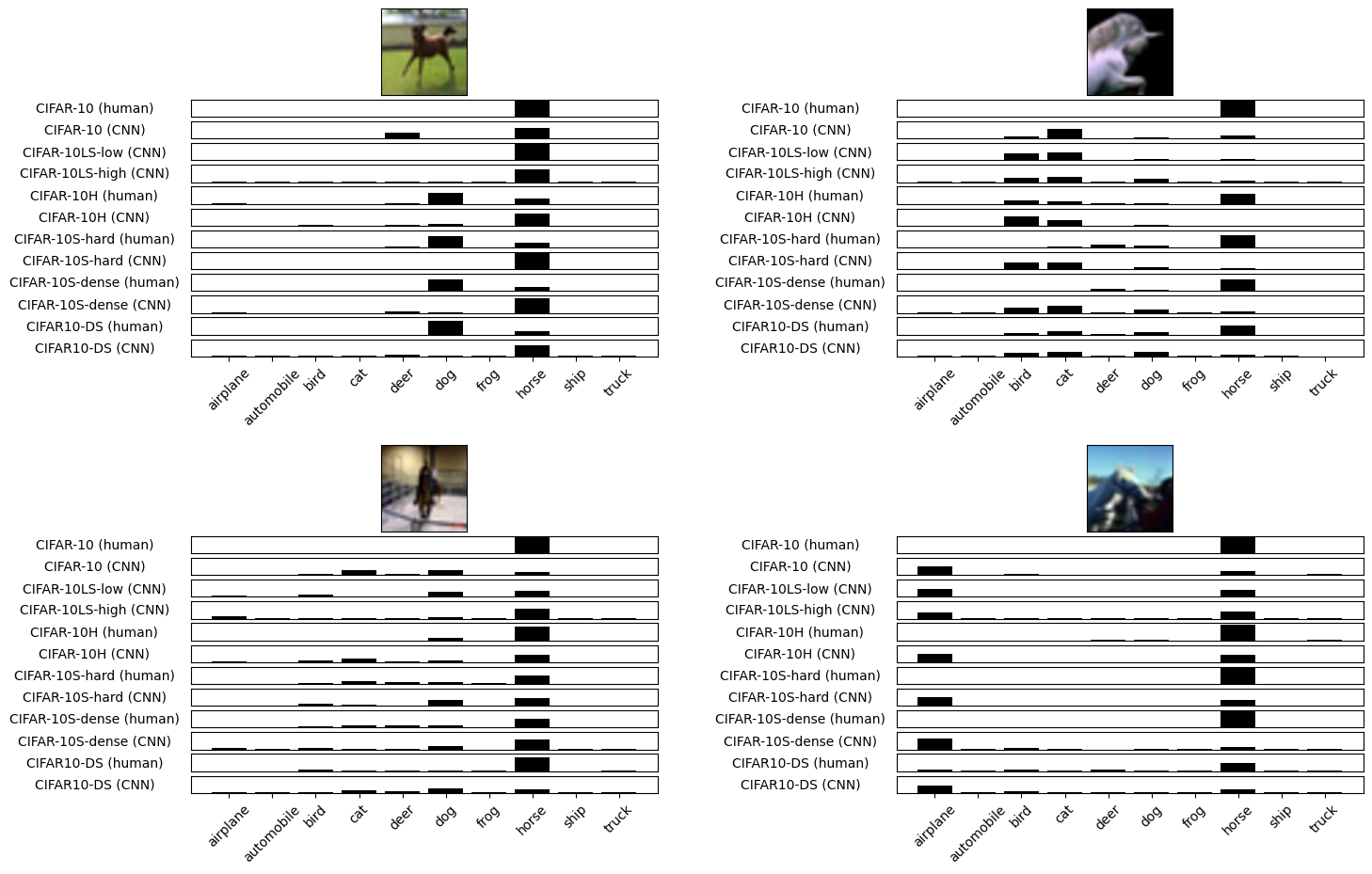}
    \caption{Representative ambiguous horse images. Top row: model agreement on high entropy image. Bottom row: maximal model disagreement.}
    \label{fig:horse}
\end{figure}

\begin{figure}[h!]
    \centering
    \includegraphics[width=0.9\textwidth]{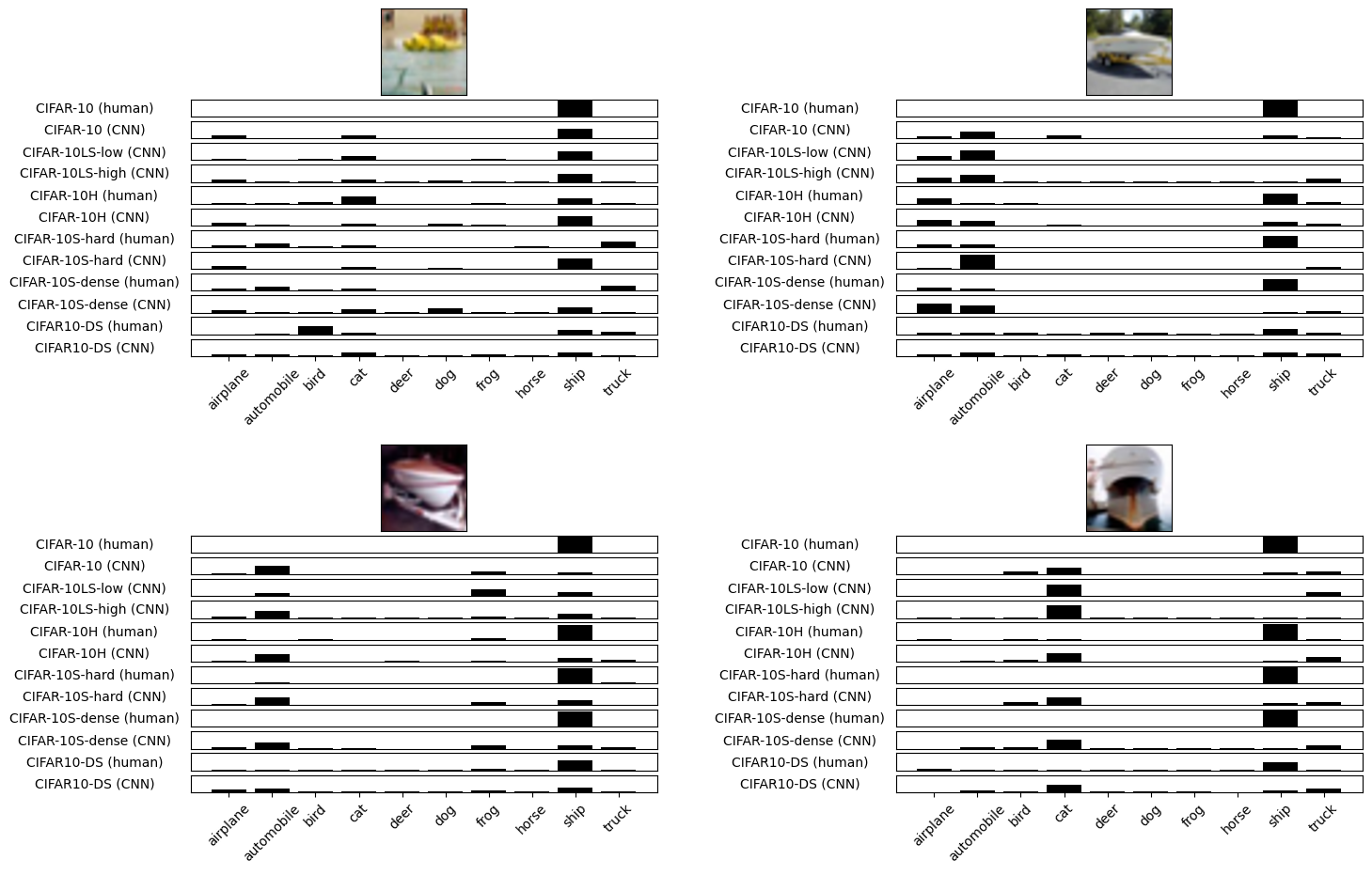}
    \caption{Representative ambiguous ship images. Top row: model agreement on high entropy image. Bottom row: maximal model disagreement.}
    \label{fig:ship}
\end{figure}

\begin{figure}[h!]
    \centering
    \includegraphics[width=0.9\textwidth]{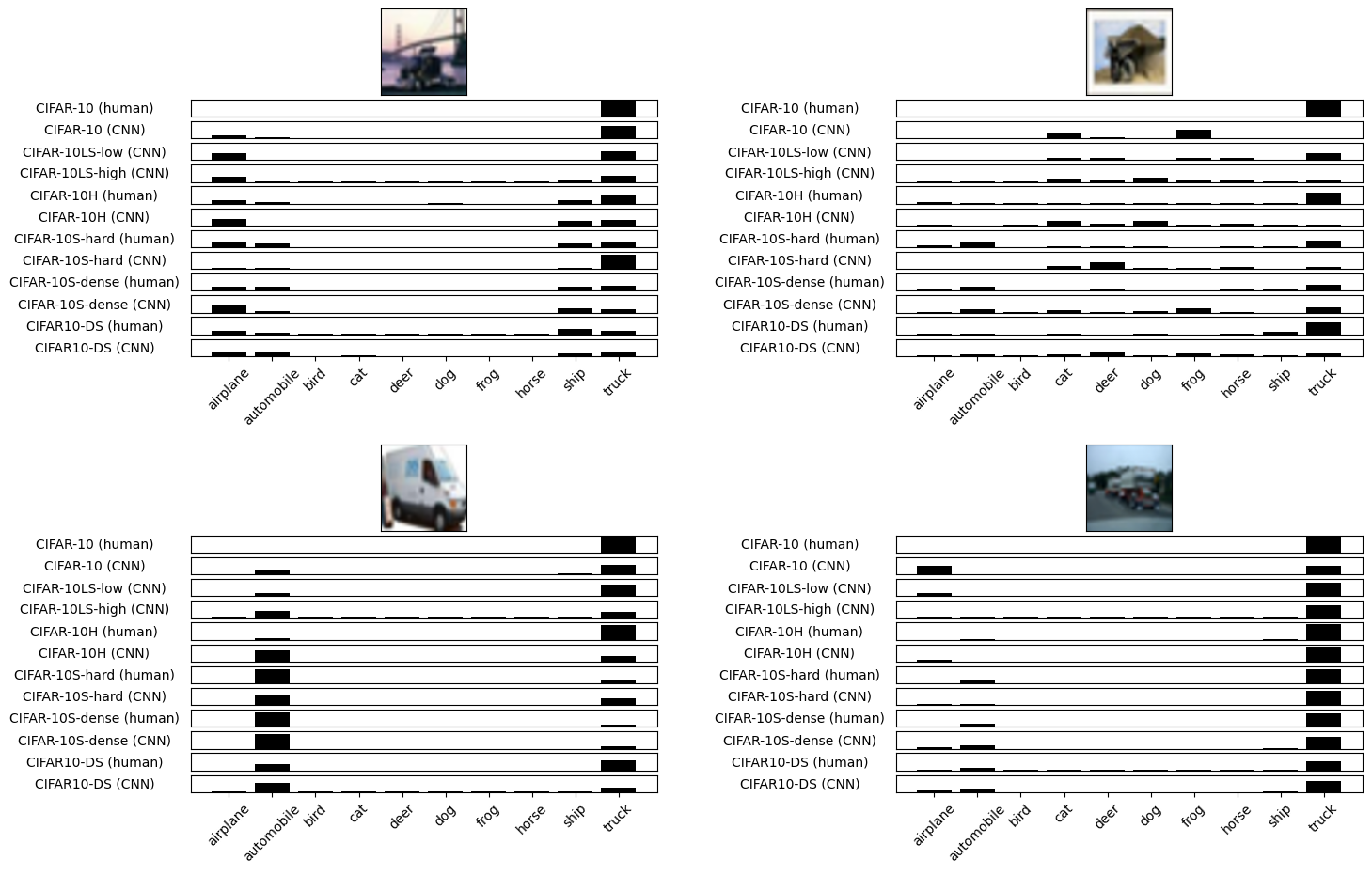}
    \caption{Representative ambiguous truck images. Top row: model agreement on high entropy image. Bottom row: maximal model disagreement.}
    \label{fig:truck}
\end{figure}

In Figures \ref{fig:plane}-\ref{fig:truck}, we present four exemplars from each class, along with the soft labels and model predictions. We see that this analysis picks out genuinely ambiguous images, with borderline cases between two classes, many classes, noisy images, and categorically uncertain images. For each image, the top row are images in which models agree on high likely entropy (average prediction entropy). The bottom row is where models maximally disagree (average symmetric relative entropy between pairwise comparisons of models).

\section{Enlarged main results figure}
\begin{sidewaysfigure}[h!]
    \centering
    \includegraphics[width=0.9\textwidth]{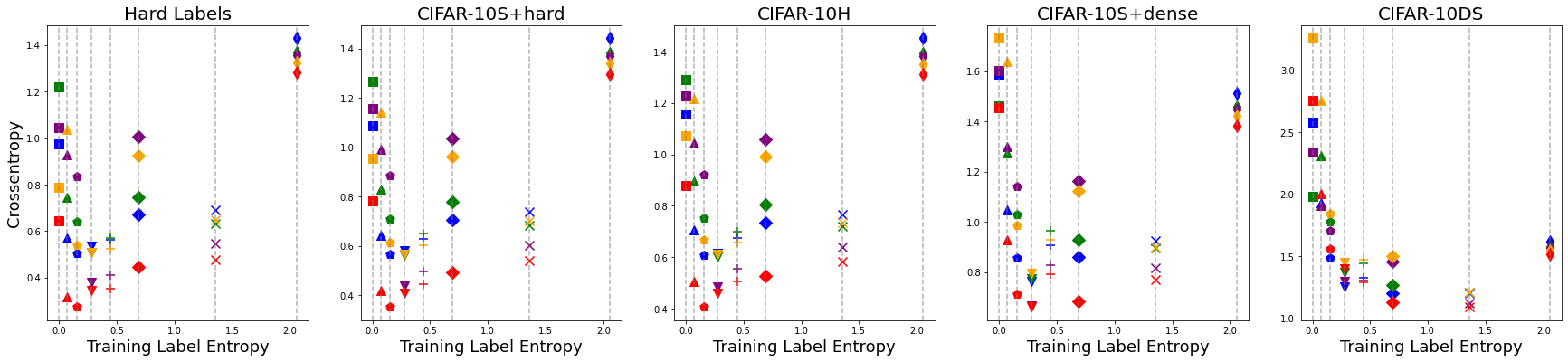}
    \includegraphics[width=0.9\textwidth]{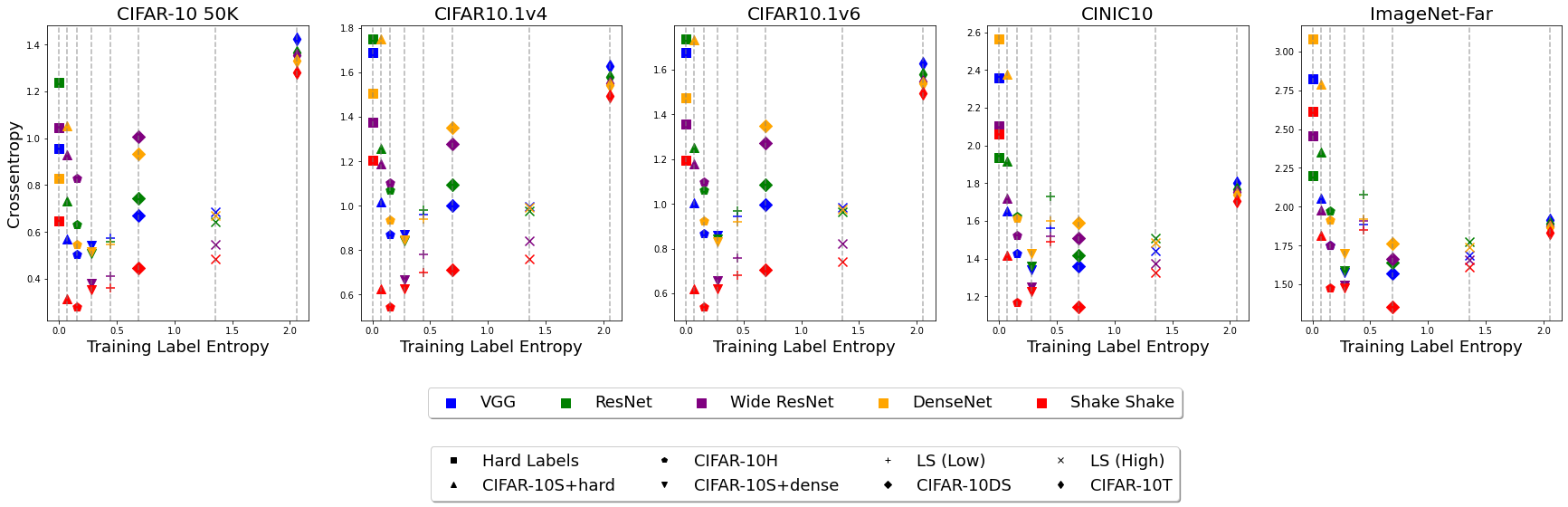}
    \vspace{-1mm}
    \caption{\textbf{Top:} Model performance on different label types at test time. \textbf{Bottom:} Generalization performance under increasing distributional shift. Each point represents the average score for a single model architecture (specified by color), trained on a particular label type (indicated via shape). Vertical lines represent points for a given label type.}
    \vspace{-3mm}
    \label{fig:genCheckLarge}
\end{sidewaysfigure}

See Figure~\ref{fig:genCheckLarge}. 

\clearpage


\end{document}